\documentclass[a4paper]{article}
\usepackage{a4wide}
\usepackage{amsmath}
\usepackage{amssymb}
\usepackage{graphicx}
\usepackage{eqnarray}
\usepackage{epstopdf}
\usepackage{subfigure}
\usepackage{algorithm}
\DeclareMathOperator*{\argmin}{\arg\!\min}

\title{OLAE-ICP: Robust and fast alignment of geometric features with the optimal linear attitude estimator\footnote{This technical report extends the preliminary description of the method mentioned in `A Modular Optimization Framework for Localization and Mapping' on the 2019 `Robotics: Science and Systems (RSS 2019)' conference.}}
\author{Jose Luis Blanco-Claraco\footnote{Department of Engineering, University of Almeria (04120), Spain, \texttt{jlblanco@ual.es}}}

\begin{document}
\maketitle

\abstract{
The problems of point-cloud registration and attitude estimation from vector observations (Wahba's problem) have widespread applications in computer vision and mobile robotics.
This work introduces a simple approach for integrating sets of geometric feature observations (points, lines, and planes) in such a way that any solution to either point-cloud registration or to Wahba's problem can be used to find the SE(3) transformation between the two sets that minimizes the corresponding cost function.
We compare the performance of three solutions: classic Horn's optimal quaternion method, Optimal Linear Attitude Estimator (OLAE) that efficiently recovers the optimal Gibbs-Rodrigues vector solving a small linear system, 
and an iterative non-linear Gauss-Newton solver.
Special care is given to explain how to overcome the Gibbs vector singularity for OLAE by using the method of sequential
rotations.
Gross outliers in point-to-point correspondences can be discarded by means of detecting transformation scale mismatches. 
The approach also allows the introduction of per-primitive relative weights, 
including an optional robust loss function that is applicable only if an initial guess for the solution is known in advance.
Experiments are presented to evaluate how the three solutions tolerate noise in the input data for different kinds of geometric primitives.
Finally, experiments with real datasets validate the suitability of the optimal alignment algorithm 
as the core of an Iterative Closest Point/Primitive (ICP) algorithm.
An open-source implementation of all the described algorithms is provided in \texttt{https://github.com/MOLAorg/mp2p\_icp}.
}

\section{Introduction}

The registration problem at hand consists of finding the transformation that best fits
a set of noisy observations of geometric primitives taken from 
the frames of references $\mathbf{x}_A \in SE(3)$ and $\mathbf{x}_B \in SE(3)$. 
That is, we are interested in finding the relative pose $\mathbf{x}_B \ominus \mathbf{x}_A$
that minimizes a particular cost function.

Fundamentally different problems arise next depending on whether the \emph{correspondences} between the two sets are known or unknown.
If they are known, an \emph{optimal transformation} algorithm, which usually has a closed-form solution, must be employed; 
otherwise, the usual approach is to apply a version of the Iterative Closest Point (ICP) (\cite{besl1992method,pomerleau2015review}) to iterate between establishing correspondences and finding the corresponding optimal transformation (using one of the algorithms of the former group) until convergence is achieved. 

Existing ICP algorithms have been reviewed extensively in the literature, 
for example, in \cite{pomerleau2015review}.
Most proposed versions deal only with point-to-point correspondences, 
or propose point-to-plane correspondences, as in \cite{segal2009generalized}.
Closed-form solution for plane to plane pairings was already proposed in the seminar work \cite{faugeras1986representation}, although line to line correspondences lacked a closed-form solution and a nonlinear iterative solution was proposed in that case.

%\cite{feldmar1996rigid} uses...

% Claims.
The present approach aims at integrating planes and lines into 
\emph{existing} optimal attitude solvers, an idea that seems 
not to be proposed in related works. Note that a fundamental limitation
of our proposal in its present form is that planes and lines 
only contribute information about the relative \emph{orientation} 
between two frames of references, thus at least one point to point pairing is 
required to completely solve the full SE(3) relative pose.

The goal of this work is, therefore, to: (i) provide a way to integrate different geometric primitives into 
the existing Horn's and OLAE solvers, and (ii) perform an experimental evaluation of their performance.

The rest of this paper is organized as follows. Section~\ref{sect:background} reviews the foundations of the related methods. 
Then, the proposed method, together with a throughout discussion of how to reliably implement OLAE is provided in Section~\ref{sect:proposed}.
Experimental results are then exposed in the next section and we finally provide some discussion and conclusions in Section~\ref{sec:conclusion}.

\section{Background}
\label{sect:background}

The algorithm in \cite{horn1987closed} to find the optimal transformation between pairs of points,
relative to their corresponding point cloud centroids, 
is very well-known in the robotics and computer vision 
communities, so it will be not further described here.

On the other hand, we have a large body of research focused on
finding the optimal relative orientation, i.e. a SO(3) transformation, 
between a set of \emph{unit vector observations}. Note that this is in contrast to \emph{relative point observations}, which may have arbitrary norms, leading to a similar, but different problem. 
The vector observation problem arises naturally in spacecraft and satellite localization, hence it has been of the 
maximum interest to the aerospace community since the original Wahba's 1965 formulation (see \cite{whaba1965least}):

\begin{equation}
\label{eq:wahba}
\mathbf{R}^\star = \argmin_{\mathbf{R} \in SO(3)} \sum_{i=1}^N w_i || \mathbf{v}^a_i -\mathbf{R} \mathbf{v}^b_i ||^2
\end{equation}

\noindent where $\mathbf{R}^\star$ is the sought optimal rotation matrix of $b$ with respect to $a$, $\mathbf{v}^a_i$ and $\mathbf{v}^b_i$ are the unit vector observations, taken in frames of reference $a$ and $b$, respectively, and $w_i$ are the relative scalar weights for each observation. 
If weights are identified as the inverse of each observation variance, 
the problem becomes a maximum likelihood estimator. 

For a survey of solutions to Eq.(\ref{eq:wahba}) that have been proposed in the literature over the years, please refer to \cite{markley1999estimate,markley2000quaternion}.

The present work focuses on a the Optimal Linear Attitude Estimator (OLAE),
as introduced in \cite{mortari2007olae}. 
OLAE allows the global estimation (noniterative, without any initial guess) of the optimal relative attitude between a set of unit vector observations.
In a sense, it is \emph{similar} to Wahba's problem (becoming equivalent only in the limit case of zero noise),
but with a different target cost function and using a 
Caley-Gibbs-Rodrigues rotation vector parameterization (\cite{bauchau2003vectorial}).
This formulation leads to the formation of a linear system, with a strictly quadratic cost function:

\begin{equation}
\label{eq:olae.org}
\mathbf{M}_m \mathbf{g} = -\mathbf{z}
\end{equation}

\noindent with $\mathbf{g}$ the sought Gibbs vector. Refer to Eq.(25) in \cite{mortari2007olae}
for the expression of $\mathbf{M}_m$; the expression for $\mathbf{z}$ will be described in \S\ref{sect:olae}.

Recently, \cite{lourakis2018efficient} proposed to use OLAE 
as an alternative to Horn's method for determining the orientation inside the general loop of ICP, 
but that work did not include any way to also handle planes or lines in the matching process.

% ================================
\section{Proposed approach}
\label{sect:proposed}
% ================================

In the following we provide details on how to build an ICP system, 
capable of handling point, line, and plane correspondences. 
At the core of the ICP algorithm we are free to choose any optimal transformation algorithm, 
hence we will describe three of them first: the well-known Horn's optimal quaternion solution in \S\ref{sect:horn},
the OLAE in \S\ref{sect:olae}, and an iterative, Gauss-Newton method in \S\ref{sect:gauss-newton} as baseline.

% ------------------------------------------
\subsection{Unifying primitives for Horn's optimal solution}
\label{sect:horn}

Let $A=\{\mathbf{a}_i\}_{i=1}^{N_a}$ and $B=\{\mathbf{b}_i\}_{i=1}^{N_b}$ 
be the sets of geometric primitives as observed from frames $\mathbf{x}_A$ and $\mathbf{x}_B$, respectively.
At this point we will assume that correspondences have been already being established between the two sets, 
hence the $i-th$ element of $A$ is believed to correspond to the $i-th$ element of $B$, 
so we let $N=N_a=N_b$ be the total number of features for simplicity of notation.

Horn's method accepts as input two sets of points, hence it is not directly suitable to handle other kind of geometric primitives. 
However, a key observation is that \cite{horn1987closed} actually estimates the transformation in three steps: first the rotation, then the scale, and finally the translation.
Rotation is decoupled by means of redefining points in \emph{local coordinates} with respect to their corresponding point cloud \emph{centroid}.
Therefore, the core of the Horn's method actually takes as input two sets of \emph{vectors},
$A_H=\{\mathbf{v}^a_i\}_{i=1}^{N}$ and $B_H=\{\mathbf{v}^b_i\}_{i=1}^{N}$, which are always assumed to represent \emph{local coordinates} of points in a point cloud.

However, nothing prevents us to include additional vectors into these sets, with any other geometric meaning.
In particular, we propose to build the sets of vectors $A_H$ and $B_H$ from the set of geometric primitives $A$ and $B$ as follows:

\begin{enumerate}
\item{If the $i$-th pair $(\mathbf{a}_i,\mathbf{b}_i)$ corresponds to a pair of points, we follow the standard procedure (\cite{horn1987closed}) and define:
	\begin{eqnarray}
		\mathbf{v}^a_i &=& \mathbf{a}_i - \bar{\mathbf{c}}_a \\
		\mathbf{v}^b_i &=& \mathbf{b}_i - \bar{\mathbf{c}}_b
	\end{eqnarray}
	\noindent where $\bar{\mathbf{c}}_{a,b}$ are the weighted centroids of all point features in  $A$ and $B$. 
	Due to the importance of the centroid in this method, we propose to remove those pairs that can be clearly 
	classified as outliers, based on a scale mismatch detector. 
	Elaborating on this later idea, please note that, if all pairings were inliers, and under the assumption of 
	zero-mean additive random noise, it should be fulfilled that $E[|\mathbf{v}^a_i|]-E[|\mathbf{v}^b_i|]=0, \forall i$, with $E[\cdot]$ the mathematical expectation. Equivalently, we could write this condition as $E[|\mathbf{v}^a_i|]/E[|\mathbf{v}^b_i|]=1$, leading to the following test for early classification of pairings as outliers:
	\begin{eqnarray}
		\label{eq:outlier.test}
	 	\frac{\max(|\mathbf{v}^a_i|,|\mathbf{v}^b_i|)}{\min(|\mathbf{v}^a_i|,|\mathbf{v}^b_i|)} - 1 < s_t 
	\end{eqnarray}
	\noindent with $s_t$ being the scale outlier detection threshold. Reasonable values for this threshold are in the range $(0.1,1.0)$. Smaller values tend to discard good pairings, while larger ones will only filter the most severe and obvious outliers. If the noise distribution of points \emph{and} outliers is known, its value can be precisely determined to fulfill with a predetermined level of confidence. A remark on how to make the detection of outliers more robust is given in the discussion of future works in \S\ref{sec:conclusion}.
}
\item{If the $i$-th pair $(\mathbf{a}_i,\mathbf{b}_i)$ corresponds to a pair of lines, we define $\mathbf{v}^a_i$ and $\mathbf{v}^b_i$ as the \emph{unit} director vectors of the corresponding lines. To solve the direction ambiguity of the director vector (i.e. any point and both $\mathbf{v}$ and $-\mathbf{v}$ define the same line), a criterion must be taken regarding the relative orientation of lines when observed from the sensor. For example, we could arbitrarily impose that director vectors should make an angle smaller than $\pi$ radians with respect to the unit vector from the sensor to the closest point of the line with respect to the sensor.
}
\item{If the $i$-th pair $(\mathbf{a}_i,\mathbf{b}_i)$ corresponds to a pair of planes, we define $\mathbf{v}^a_i$ and $\mathbf{v}^b_i$ as the \emph{unit} normal vectors of the corresponding planes. Again, ambiguity should be addressed to enforce that the same plane, when observed from two different view points, has a vector that points in the same direction. A natural choice in this case is to select the outwards-pointing direction of measured surfaces.
}
\end{enumerate}

Once we have all the vectors stacked into the lists $A_H$ and $B_H$, including their pairwise relative weights $w_i$, 
we follow the standard method described in \cite{horn1987closed} to recover the optimal SO(3) transformation, then use the point cloud centroids (excluding outliers) to solve for optimal translation.

% ------------------------------------------
\subsection{Unifying primitives for OLAE}
\label{sect:olae}

Just like in the former section, we start with the input sets 
$A=\{\mathbf{a}_i\}_{i=1}^{N}$ and $B=\{\mathbf{b}_i\}_{i=1}^{N}$. 
Conversely to Horn's method, vector-based attitude estimator as OLAE or any other solution to the Wahba's problem, 
accepts two sets of \emph{unit vectors}, $A_O$ and $B_O$, as input.
Again, we propose to include other entities, like points, by converting them appropriately.
We can build the sets of unit vectors $A_O=\{\mathbf{v}^a_i\}_{i=1}^{N}$ and $B_O=\{\mathbf{v}^b_i\}_{i=1}^{N}$ from the set of geometric primitives $A$ and $B$ as follows:

\begin{enumerate}
	\item{For pairs of points $(\mathbf{a}_i,\mathbf{b}_i)$, we start following the same procedure than described for the Horn's method above, 
		then we \emph{normalize} the local coordinates of the points with respect to the centroids $\bar{\mathbf{c}}_{a,b}$, that is:
		\begin{eqnarray}
		\label{eq:olae.vec.norm}
		\bar{\mathbf{v}}^a_i &=& \mathbf{a}_i - \bar{\mathbf{c}}_a \quad \quad   \mathbf{v}^a_i = \frac{\bar{\mathbf{v}}^a_i}{|\bar{\mathbf{v}}^a_i|} \\
		\bar{\mathbf{v}}^b_i &=& \mathbf{b}_i - \bar{\mathbf{c}}_b \quad \quad   \mathbf{v}^b_i = \frac{\bar{\mathbf{v}}^b_i}{|\bar{\mathbf{v}}^b_i|} 
		\end{eqnarray}
		\noindent Note that this simple transformation enables attitude estimation algorithms to also cope with point correspondences. 
		Outliers can be also detected in this case using Eq.~(\ref{eq:outlier.test}) without modifications.
	}
	\item{For line or plane correspondences, the unit director and normal vectors, respectively, already are unit vectors (the natural input to OLAE), 
		so no further actions are required.
	}
\end{enumerate}

Once we have all observations stacked into the lists of unit vectors
$A_O$ and $B_O$, and we are given a vector of relative weights $w_i$, 
we can find the $3 \times 3$ attitude profile matrix $\mathbf{B}$ and the $\mathbf{z}$ vector in Eq.~(\ref{eq:olae.org}) as:

\begin{eqnarray}
\mathbf{B} &=& \sum_{i=1}^N w_i \mathbf{v}^b_i (\mathbf{v}^a_i)^\top \\
\mathbf{v} &=& - \sum_{i=1}^N w_i \mathbf{v}^b_i  \times \mathbf{v}^a_i
\end{eqnarray}

\noindent from which can be approximate $\mathbf{M}_m$ in Eq.~(\ref{eq:olae.org})
as $\mathbf{M}_m \approx \mathbf{M}_w$ (which is more convenient for reasons that will be clear when dealing with the sequential rotation method), and where $\mathbf{M}_w$ is (see \cite{mortari2007olae}):

\begin{equation}
\label{eq:olae.Mw}
\mathbf{M}_w = \begin{bmatrix}
S_{11} - p   & S_{12} & S_{13} \\
S_{12}    & S_{22} - p & S_{23} \\
S_{13}   & S_{23} & S_{33} -p 
\end{bmatrix}
\end{equation}

\noindent where it has been used:

\begin{eqnarray}
\mathbf{S} &=& \mathbf{B} + \mathbf{B}^\top \\
p&=& tr(B)+1 \\
m&=& tr(B)-1
\end{eqnarray}

\noindent leading to the linear system of equations:

\begin{eqnarray}
\label{eq:olae.final.Mw}
\mathbf{M}_w &=& \mathbf{g} \mathbf{z}
\end{eqnarray}

\noindent from which the optimal rotation can be solved for as 
the Gibbs vector $\mathbf{g} = (g_x, g_y, g_z)$. 
In order to recover a quaternion $(q_r, q_x,q_y,q_z)$ from $\mathbf{g}$, we can use:

\begin{eqnarray}
q_r &=& \frac{1}{\sqrt{1+g_x^2+g_y^2+g_z^2}} \\
q_x &=& q_r g_x \\
q_y &=& q_r g_y \\
q_z &=& q_r g_z
\end{eqnarray}

The only edge case that remains to be dealt with is the singularity of 
the Gibbs vector representation when representing a rotation of 180 degrees.
In fact, the accuracy of the solution of the linear system in Eq.~(\ref{eq:olae.final.Mw}) may in theory be compromised when working near the singularity. In practice, for reasonable noise levels in the input data, the solution is robust even with rotations of 179 degrees.

Nevertheless, in order to work near the optimal conditions of OLAE, we propose to evaluate \emph{four} different solutions: one for the unmodified linear system in Eq.~(\ref{eq:olae.final.Mw}), and three for systems that have undergone a \emph{rotation} around each one of the axes (x,y,z).
The approximation  $\mathbf{M}_m \approx \mathbf{M}_w$ 
reveals useful here, since the rotated $\mathbf{M}_w$ matrices 
can be evaluated in closed form from $\mathbf{B}$, avoiding the need to rotate all the input vectors and going through the summation again.
This method, originally described in \cite{shuster1981attitude}, 
can be found in Eqs.(35)-(37) in \cite{mortari2007olae}

We propose to evaluate the determinant of the matrix of coefficients 
of the linear system (i.e. $\mathbf{M}_w$ for the unrotated case, $\mathbf{M}_w^{x,y,z}$ for the rotated systems) and use the system with the largest absolute value, since it will ensure that the system has a full rank of 3.

%Upper bound limit phi: \cite{markley1999estimate}.

% ------------------------------------------
\subsection{Gauss-Newton iterative solver}
\label{sect:gauss-newton}

In order to validate the implementation of the multi-primitive optimal transformation algorithms based on the Horn's method and OLAE, 
we also implemented a baseline method based on a non-linear, iterative solver. 
It support the same kind of pairings than the aforementioned methods, plus additional ones, e.g. point-to-plane. 
The interested reader is referred to the source code for details on the cost functions. 
Closed-form Jacobians have been found for each kind of pairing by applying the chain rule 
to the corresponding target function and making use of the expressions in 
\cite{blanco2010tutorial} 
to solve for SE(3) on-manifold increments at each iteration.

% ------------------------------------------
\subsection{Complete ICP algorithm}
\label{sect:complete.icp}

Once the optimal transformation estimators have been defined, 
we can build a complete multi-primitive alignment algorithm 
by iterating between selecting closest correspondences between the two 
maps, and finding the optimal transformation that raises from the selected pairings.
We use the nearest neighbor criterion for pairing points, using KD-trees for 
efficiency. Planes are paired by looking for the nearest centroids in the reference map whose normal 
vector makes an angle below a certain threshold in the map to be aligned.
A robust kernel is optionally applied by multiplying the relative weights of each pair with a robust loss function, i.e. $w_i' = w_i f(\mathbf{v}^a_i,\mathbf{v}^b_i)$.
The interested reader is referred to the source code for further details.

\begin{figure*}
	\centering
	\subfigure[SO(3) error ]{\includegraphics[width=0.49\columnwidth]{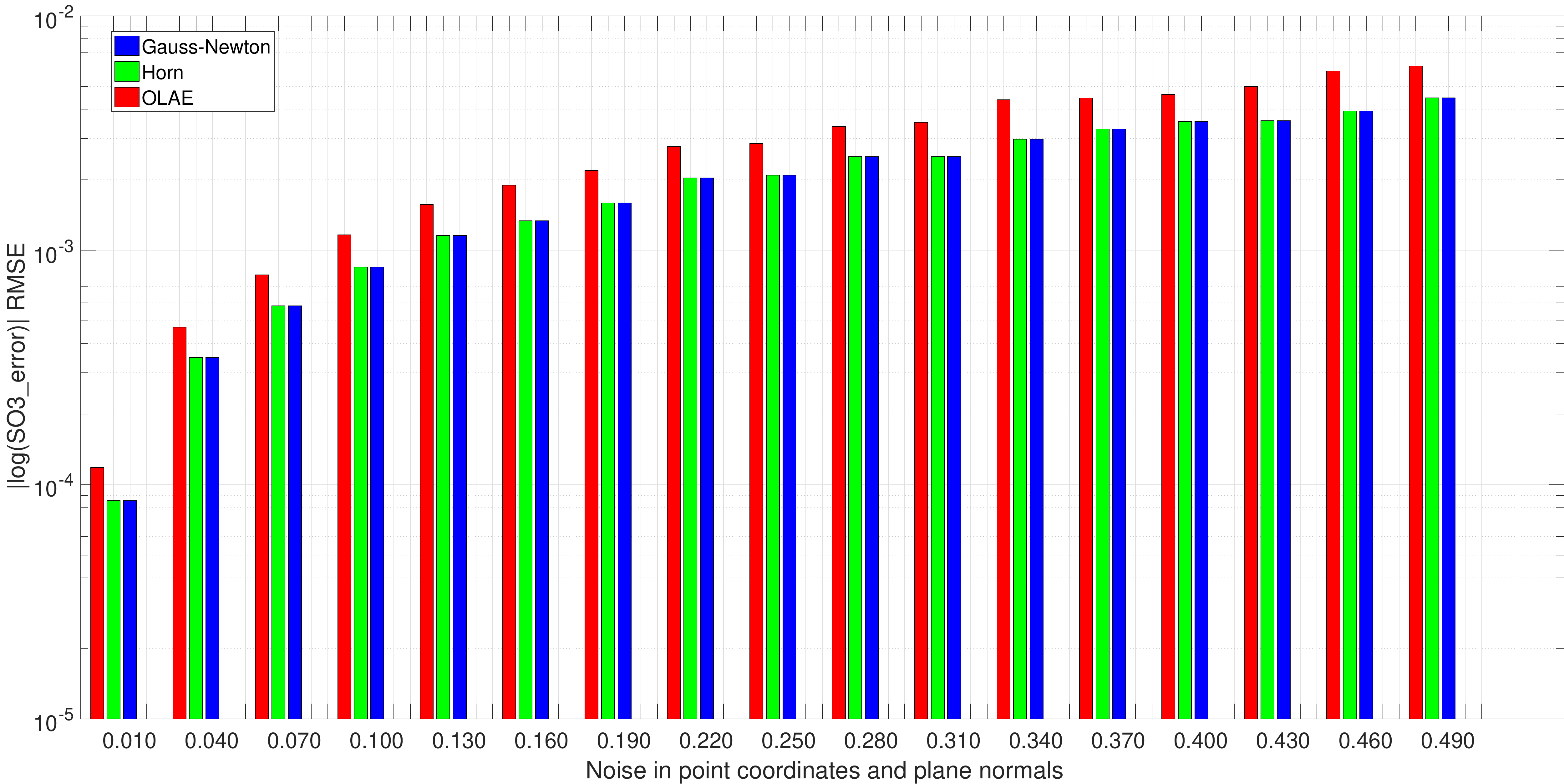}}
	\subfigure[Translation error]{\includegraphics[width=0.49\columnwidth]{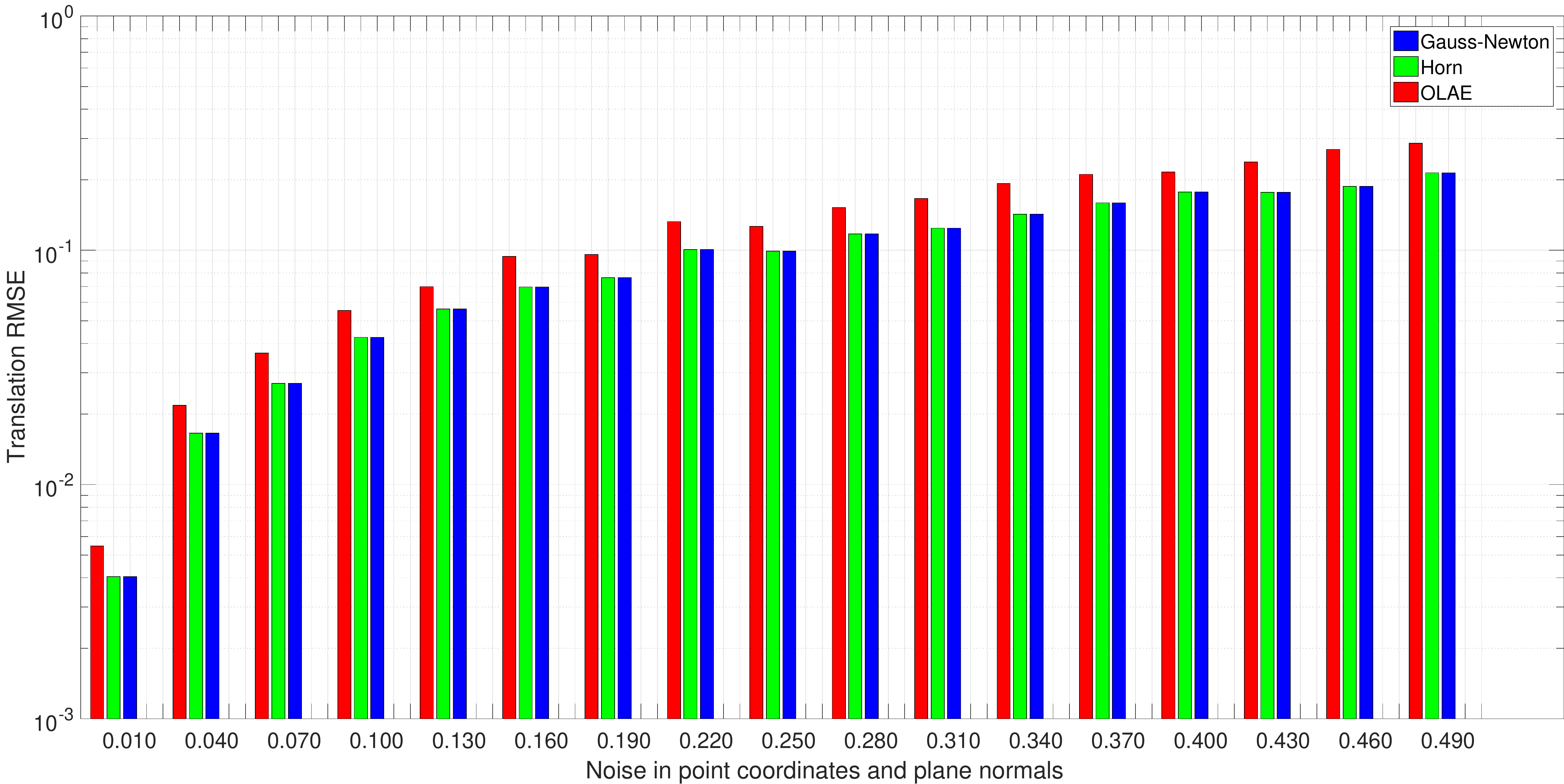}}
	\\
	\subfigure[SO(3) error for OLAE only]{\includegraphics[width=0.49\columnwidth]{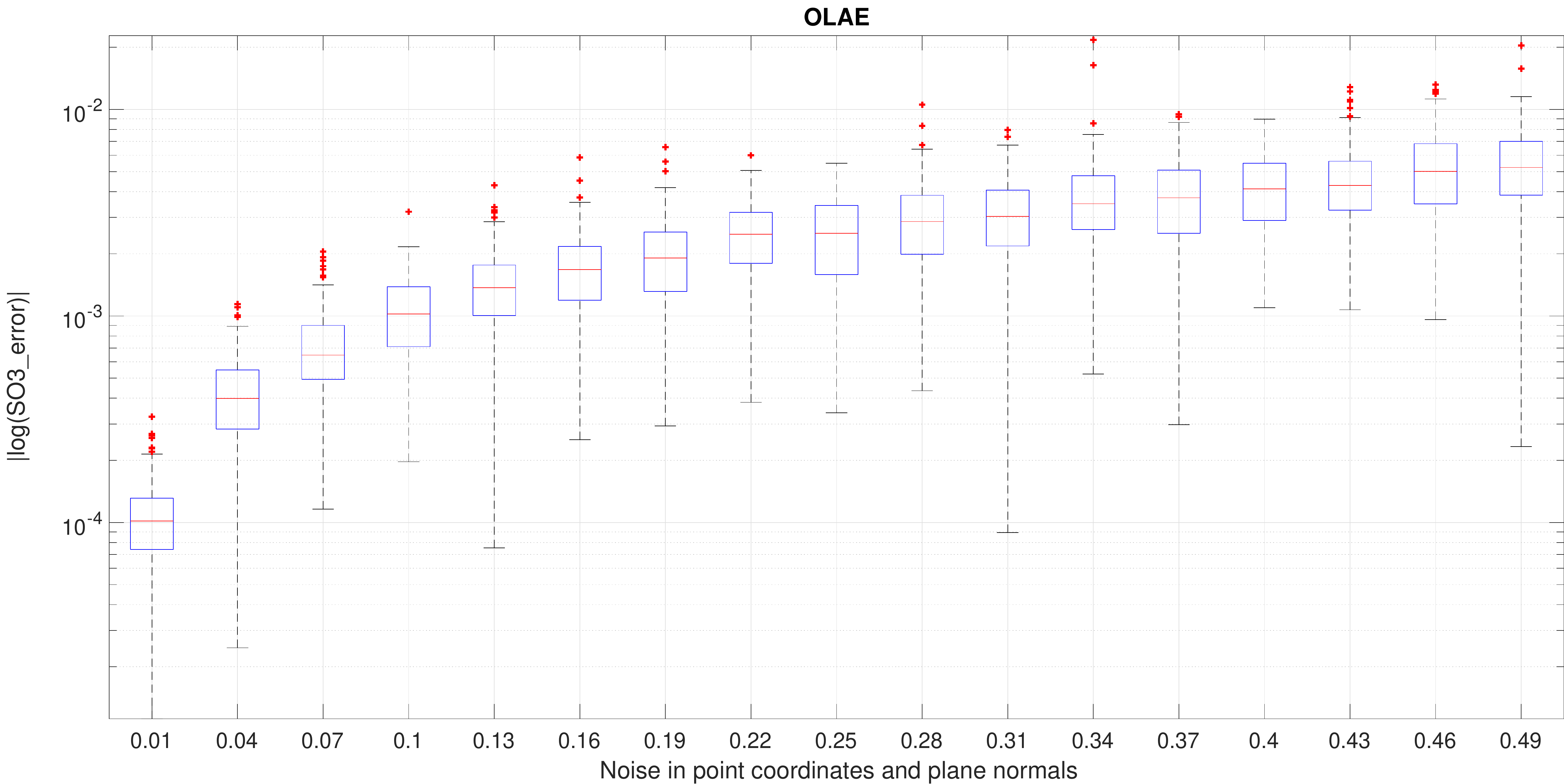}}
	\subfigure[Translation error for OLAE only]{\includegraphics[width=0.49\columnwidth]{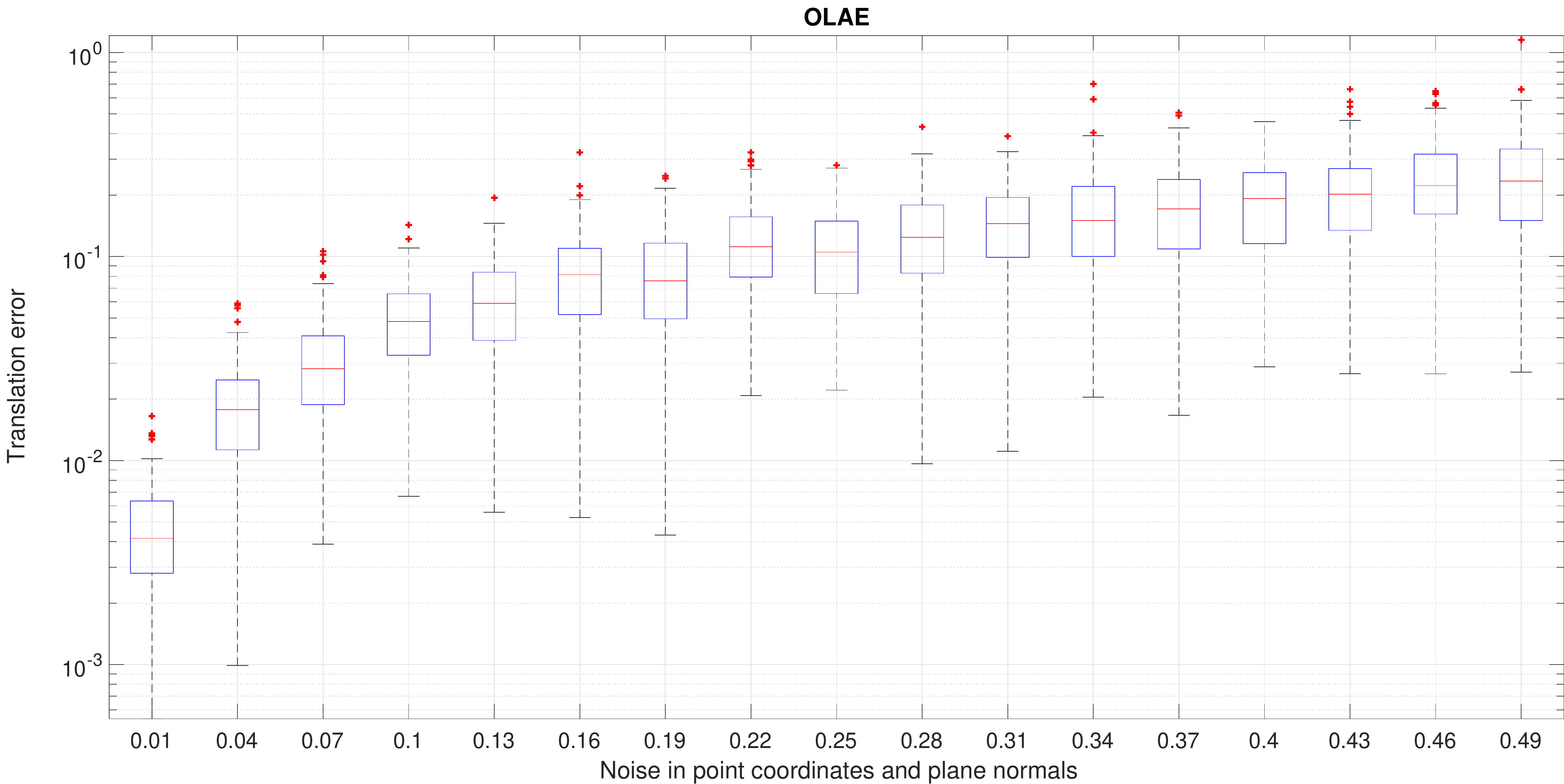}}
	\\
	\subfigure[SO(3) error for Horn's only]{\includegraphics[width=0.49\columnwidth]{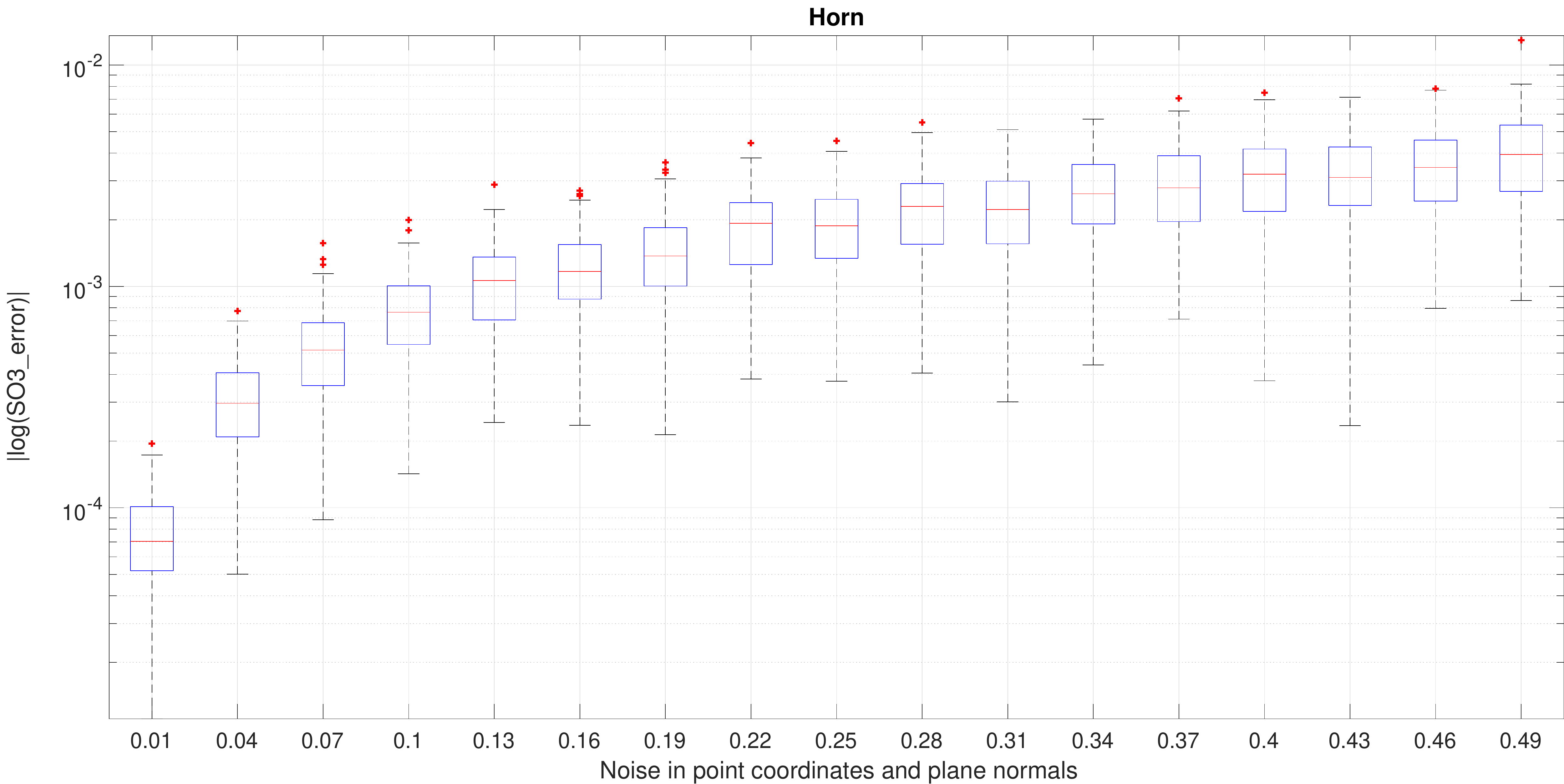}}
	\subfigure[Translation error for OLAE only]{\includegraphics[width=0.49\columnwidth]{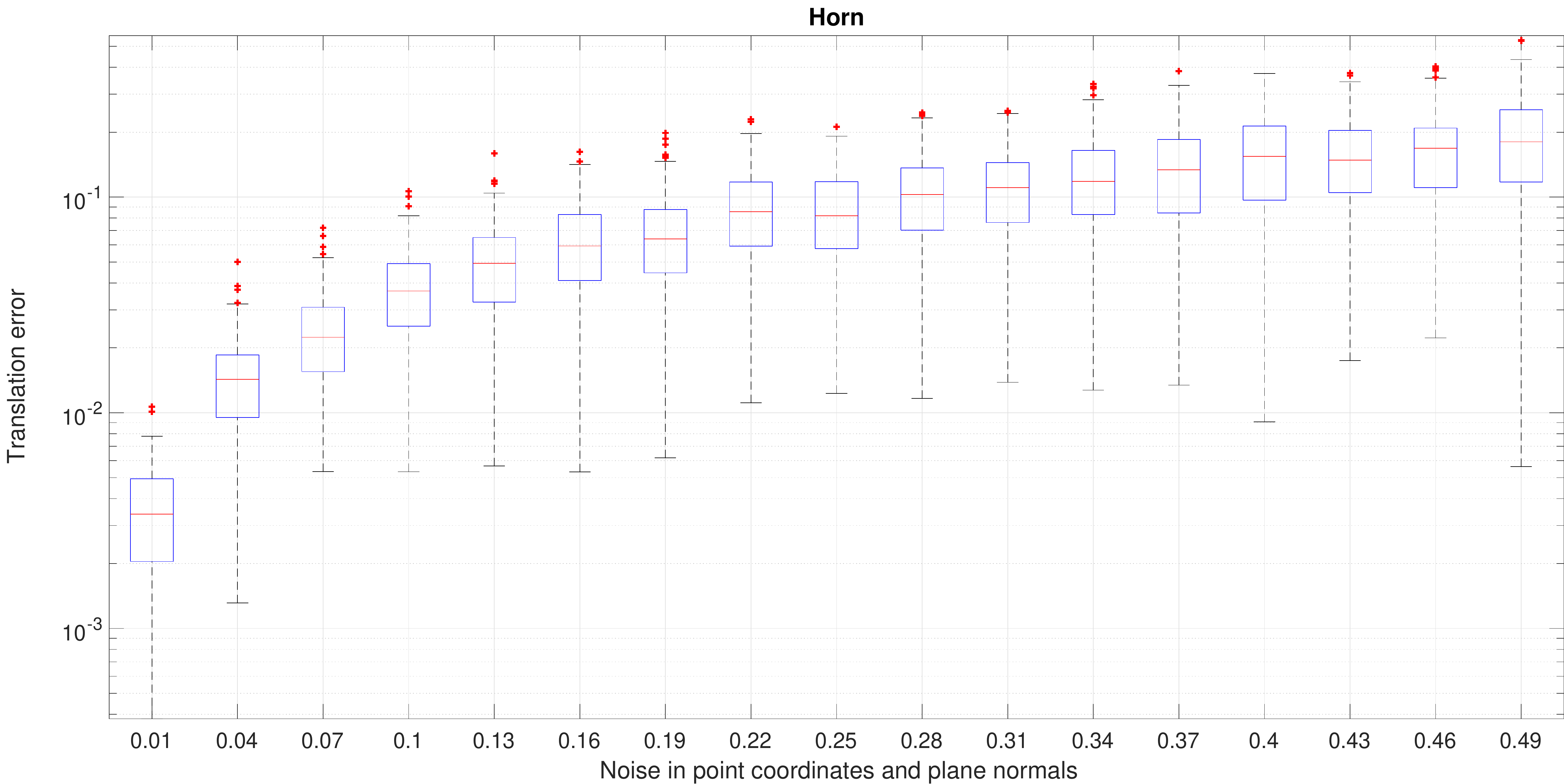}}
	\\
	\subfigure[SO(3) error for Horn's only]{\includegraphics[width=0.49\columnwidth]{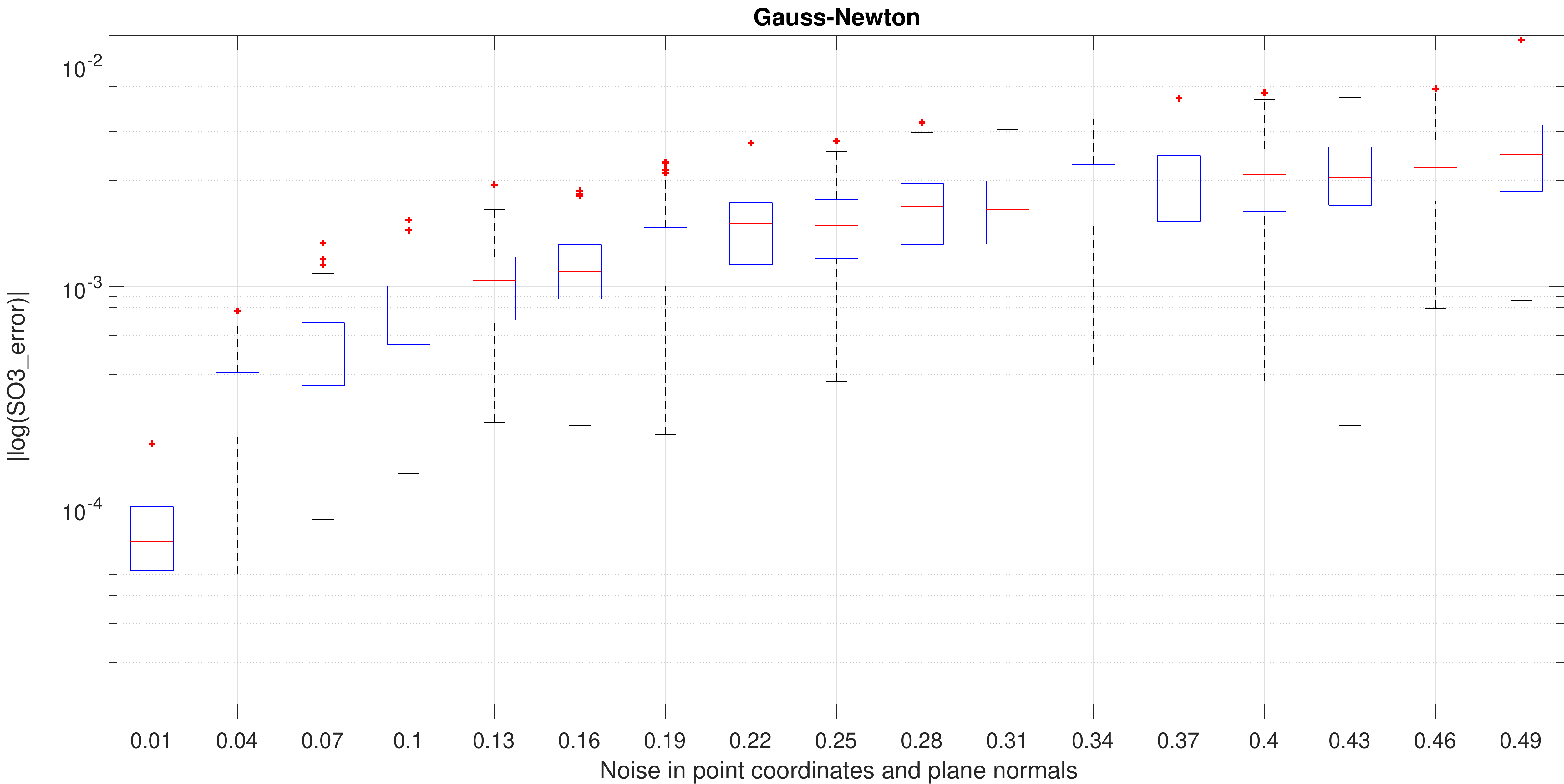}}
	\subfigure[Translation error for OLAE only]{\includegraphics[width=0.49\columnwidth]{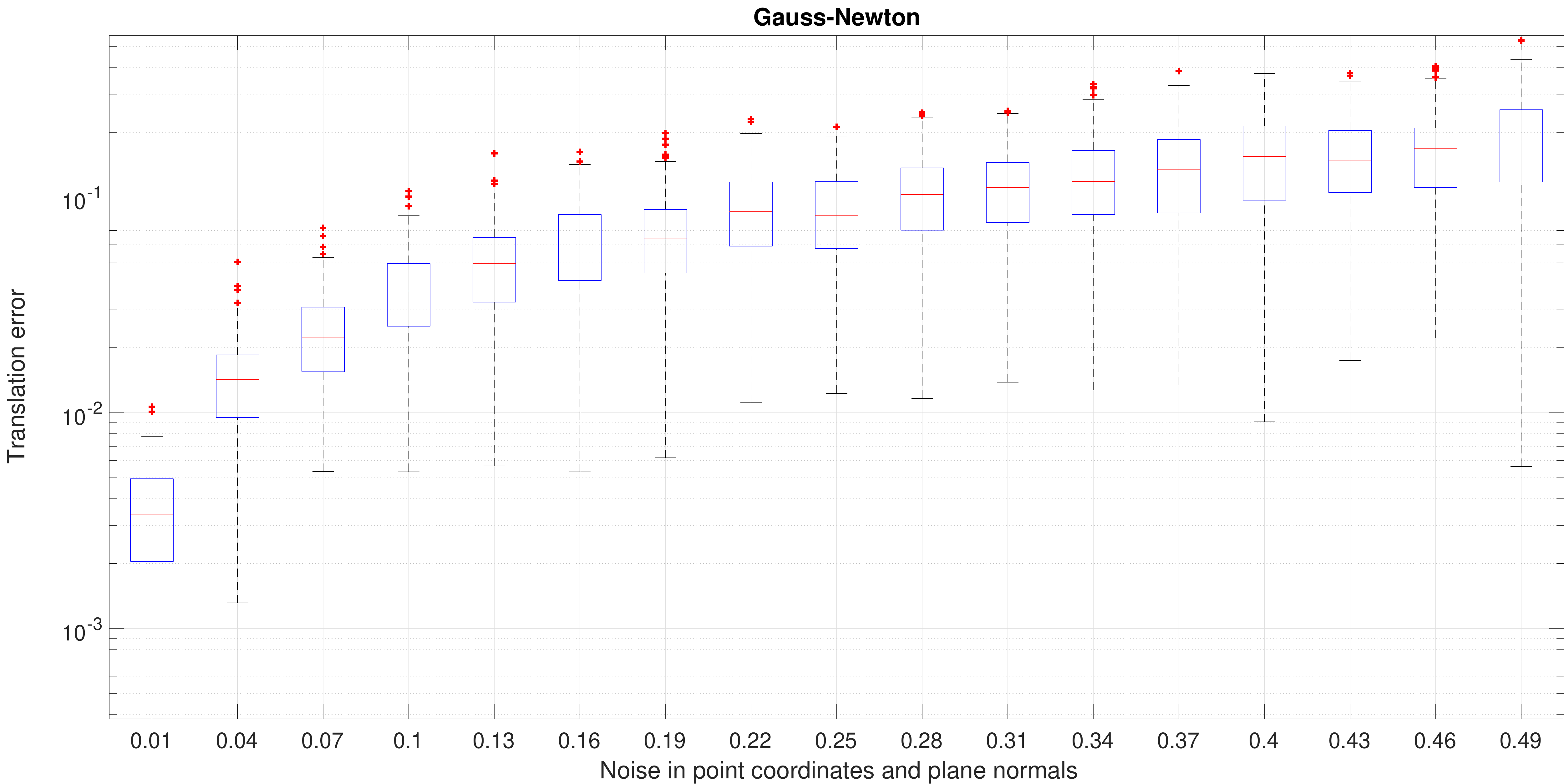}}
	\caption{Errors for the optimal transformation estimated by three different methods (OLAE, Horn's optimal quaternion, Gauss-Newton)
		for 100 point-to-point correspondences. The horizontal axis represents the standard deviation of the points additive isotropic Gaussian noise.}
	\label{fig:results.pt100.pl0}
\end{figure*}

\begin{figure*}
	\centering
	\subfigure[SO(3) error ]{\includegraphics[width=0.49\columnwidth]{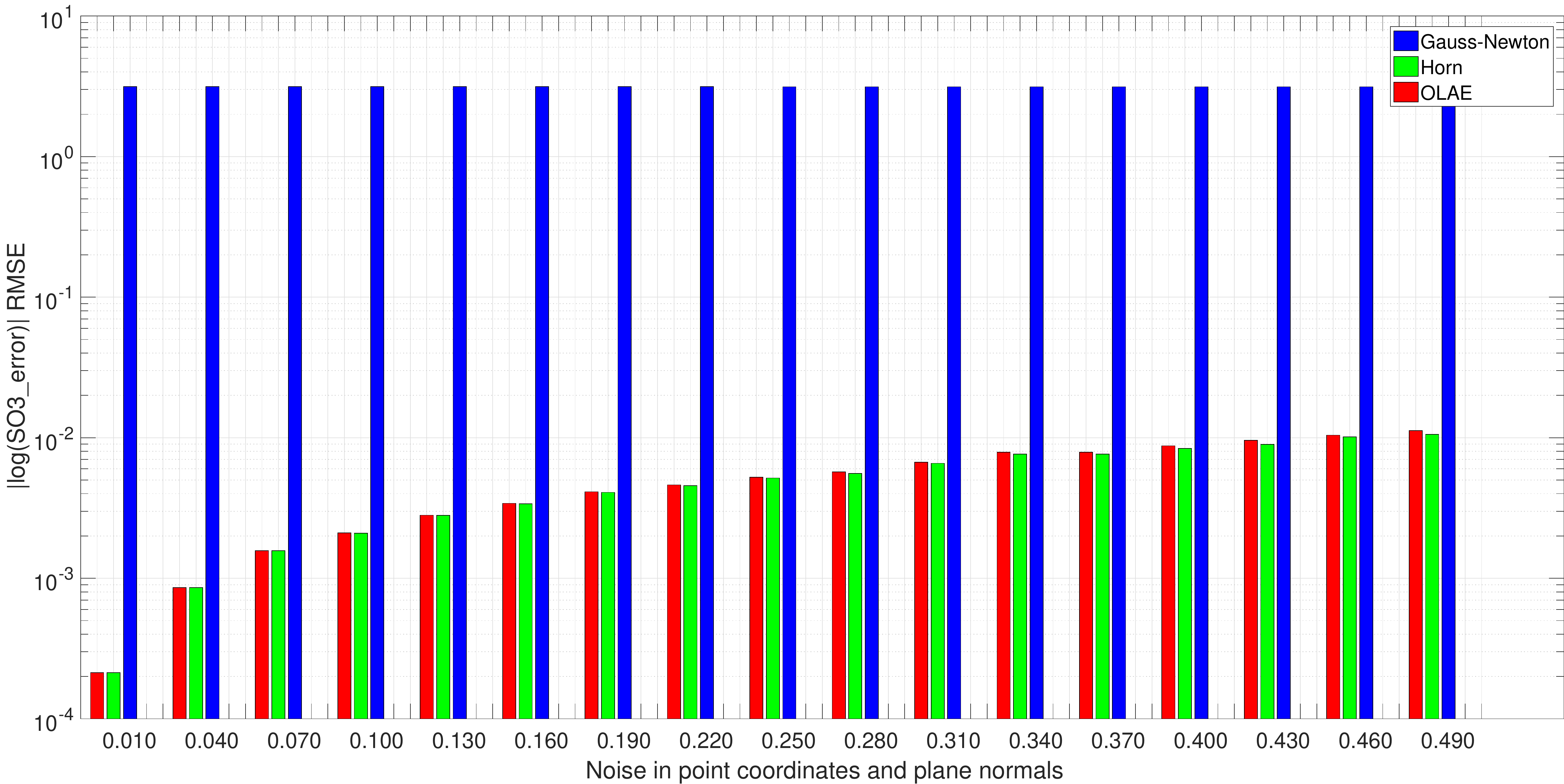}}
	\subfigure[Translation error]{\includegraphics[width=0.49\columnwidth]{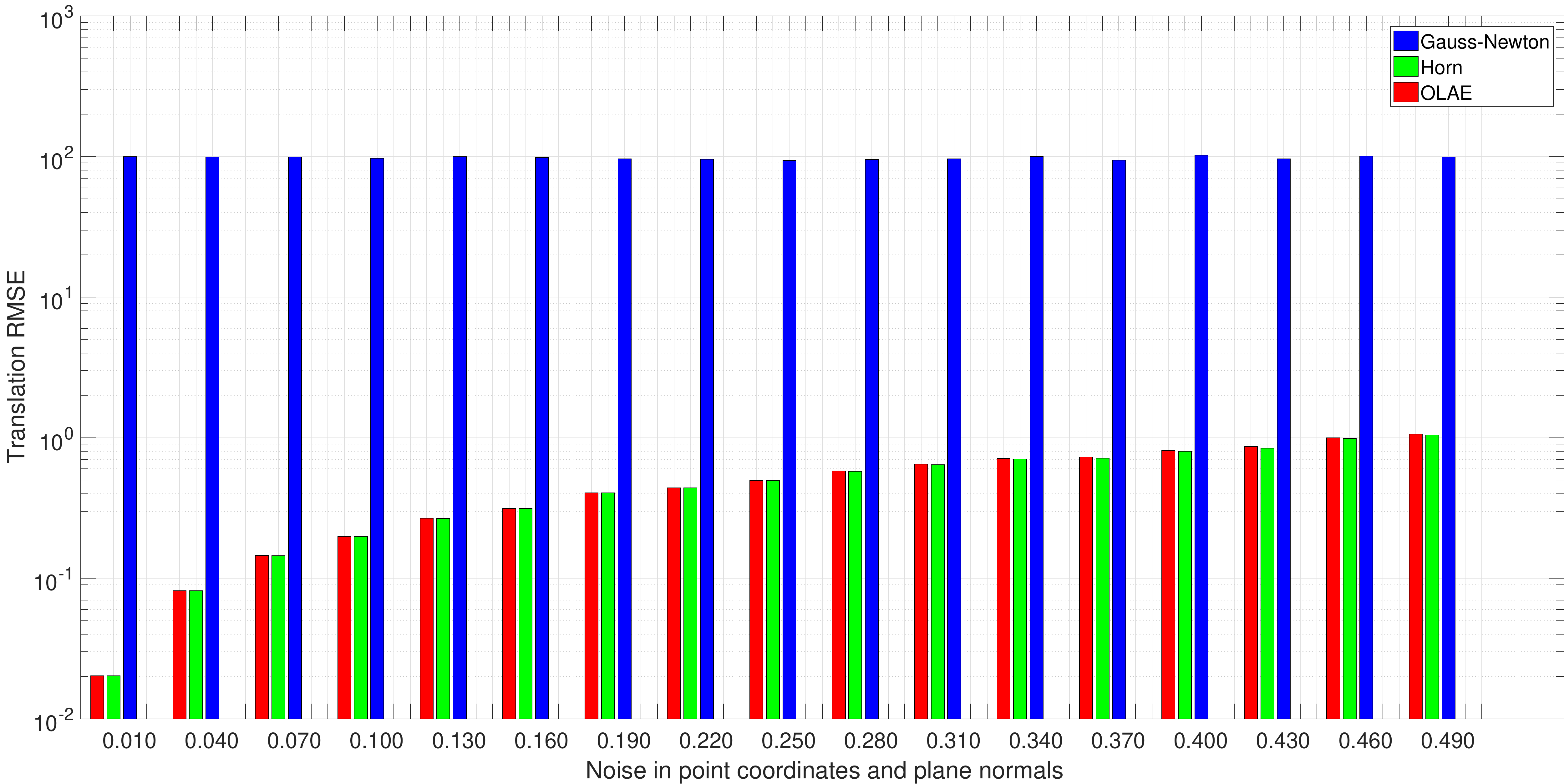}}
	\\
	\subfigure[SO(3) error for OLAE only]{\includegraphics[width=0.49\columnwidth]{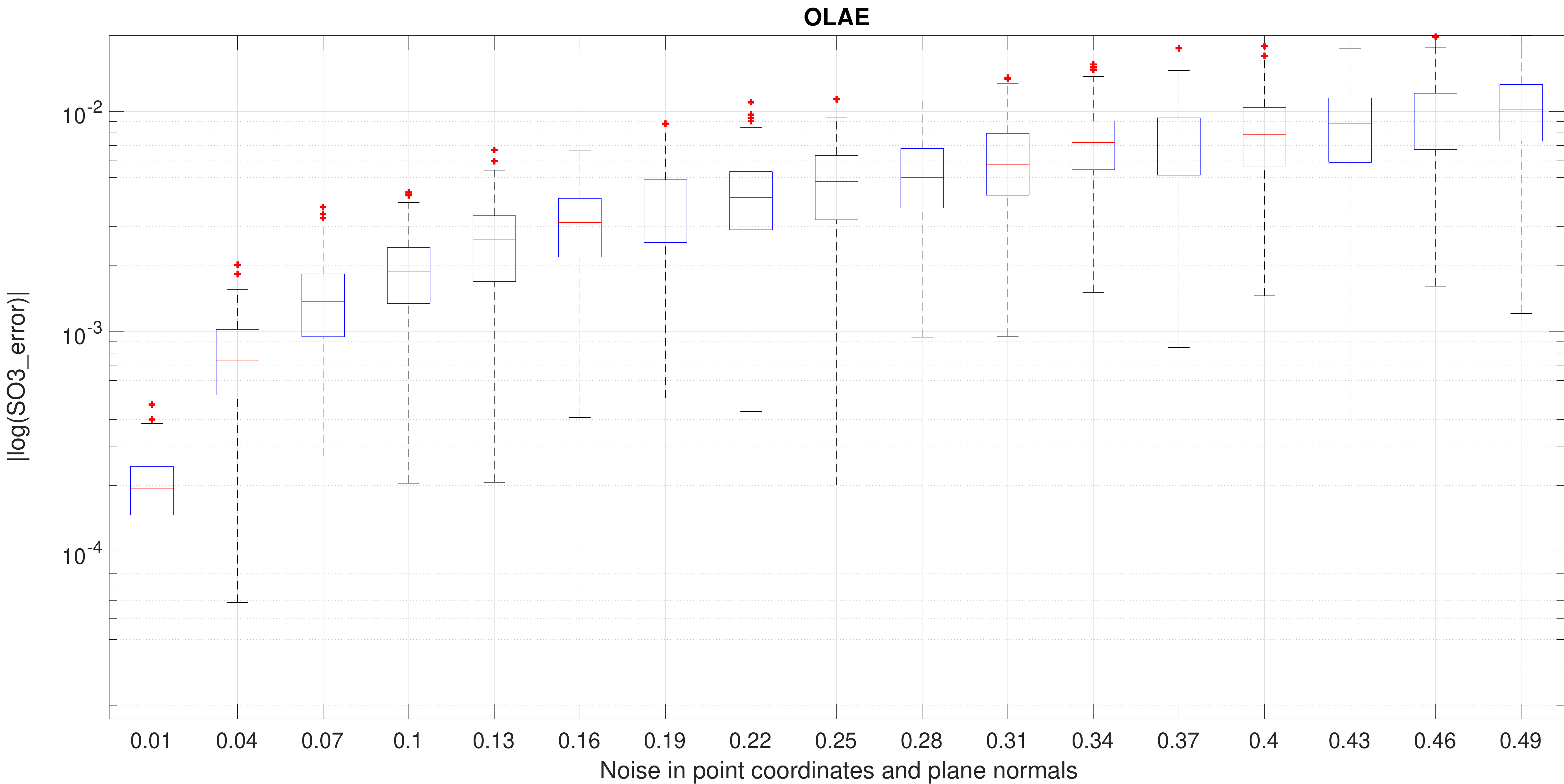}}
	\subfigure[Translation error for OLAE only]{\includegraphics[width=0.49\columnwidth]{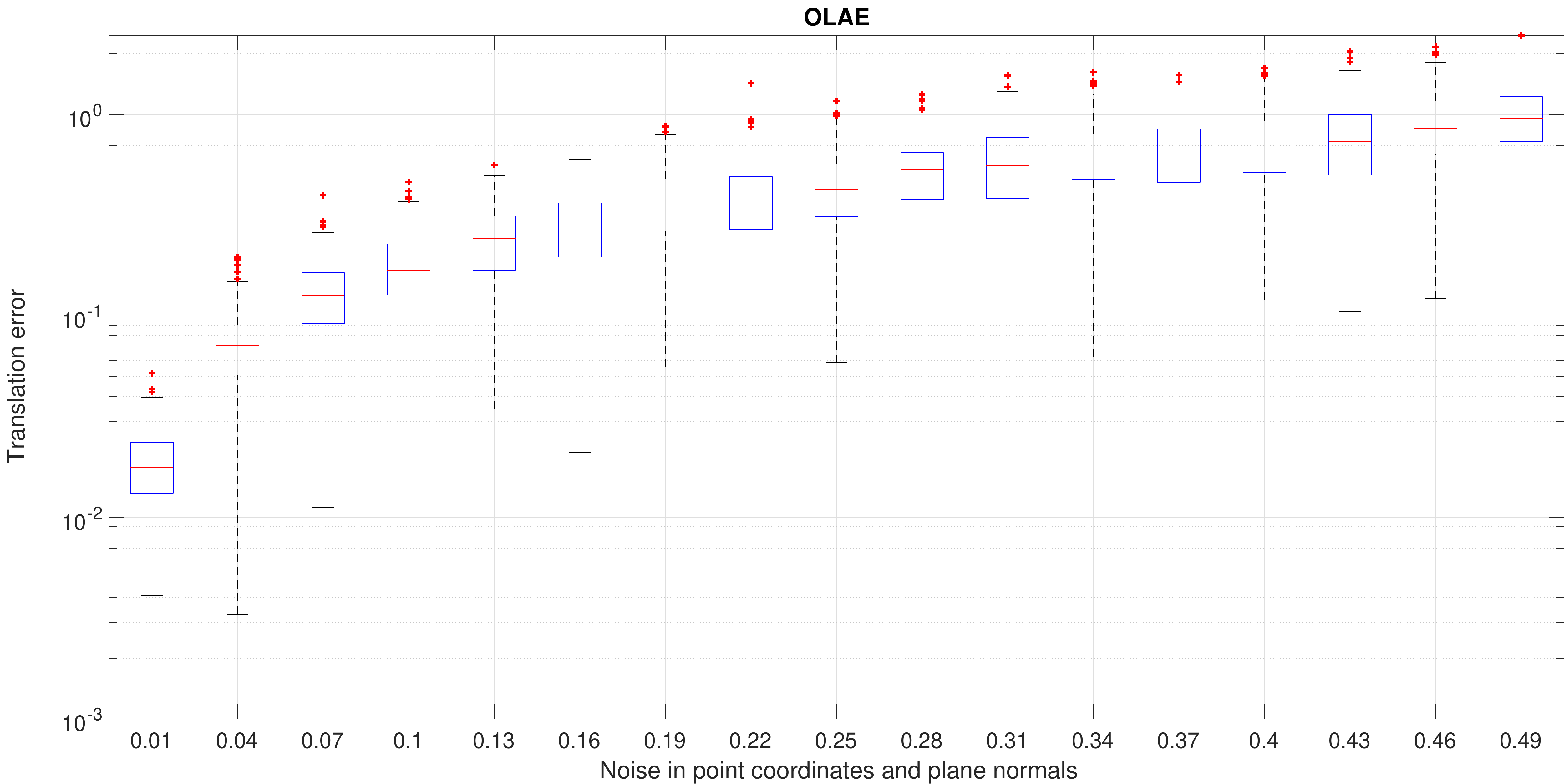}}
	\\
	\subfigure[SO(3) error for Horn's only]{\includegraphics[width=0.49\columnwidth]{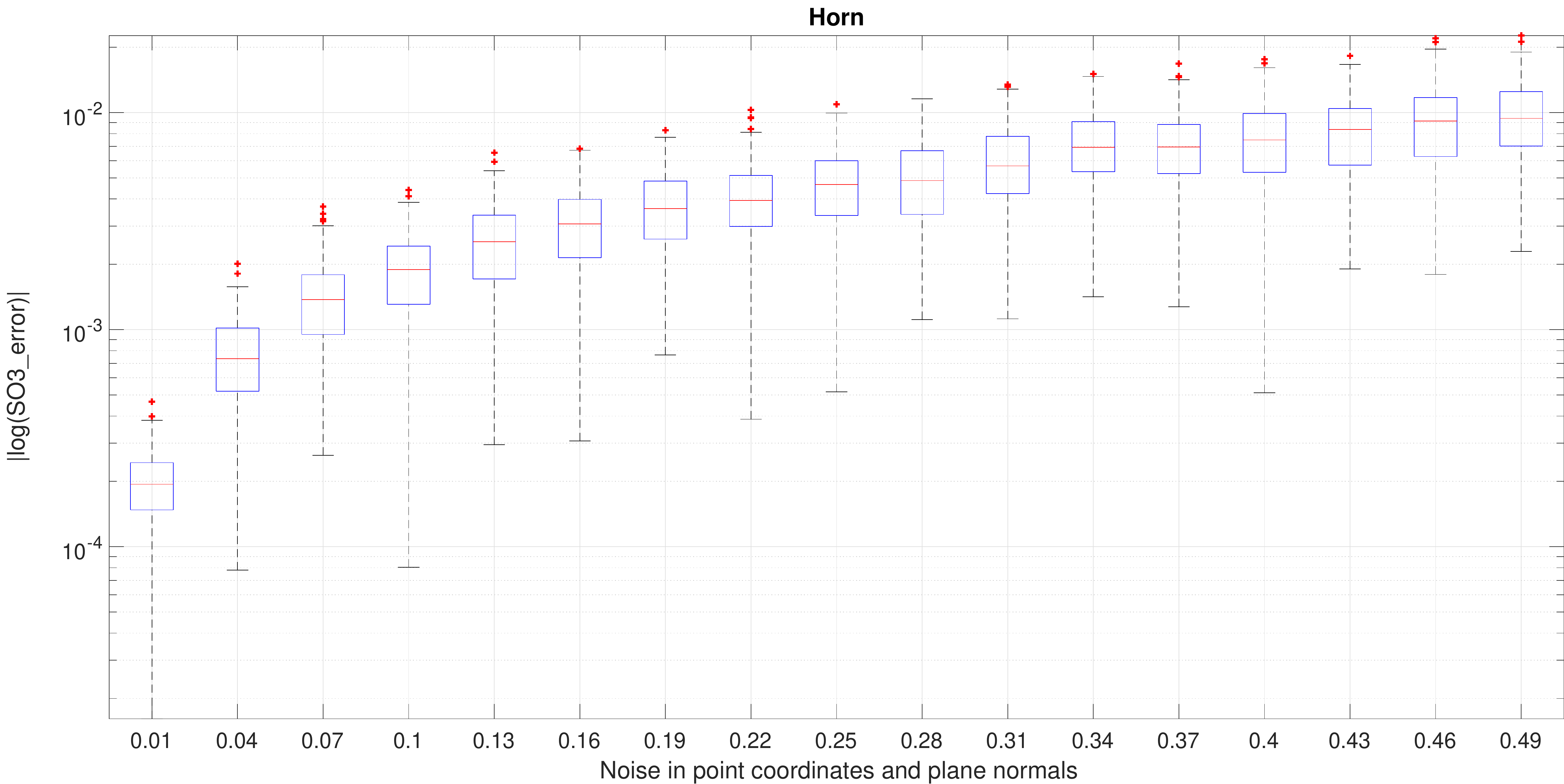}}
	\subfigure[Translation error for OLAE only]{\includegraphics[width=0.49\columnwidth]{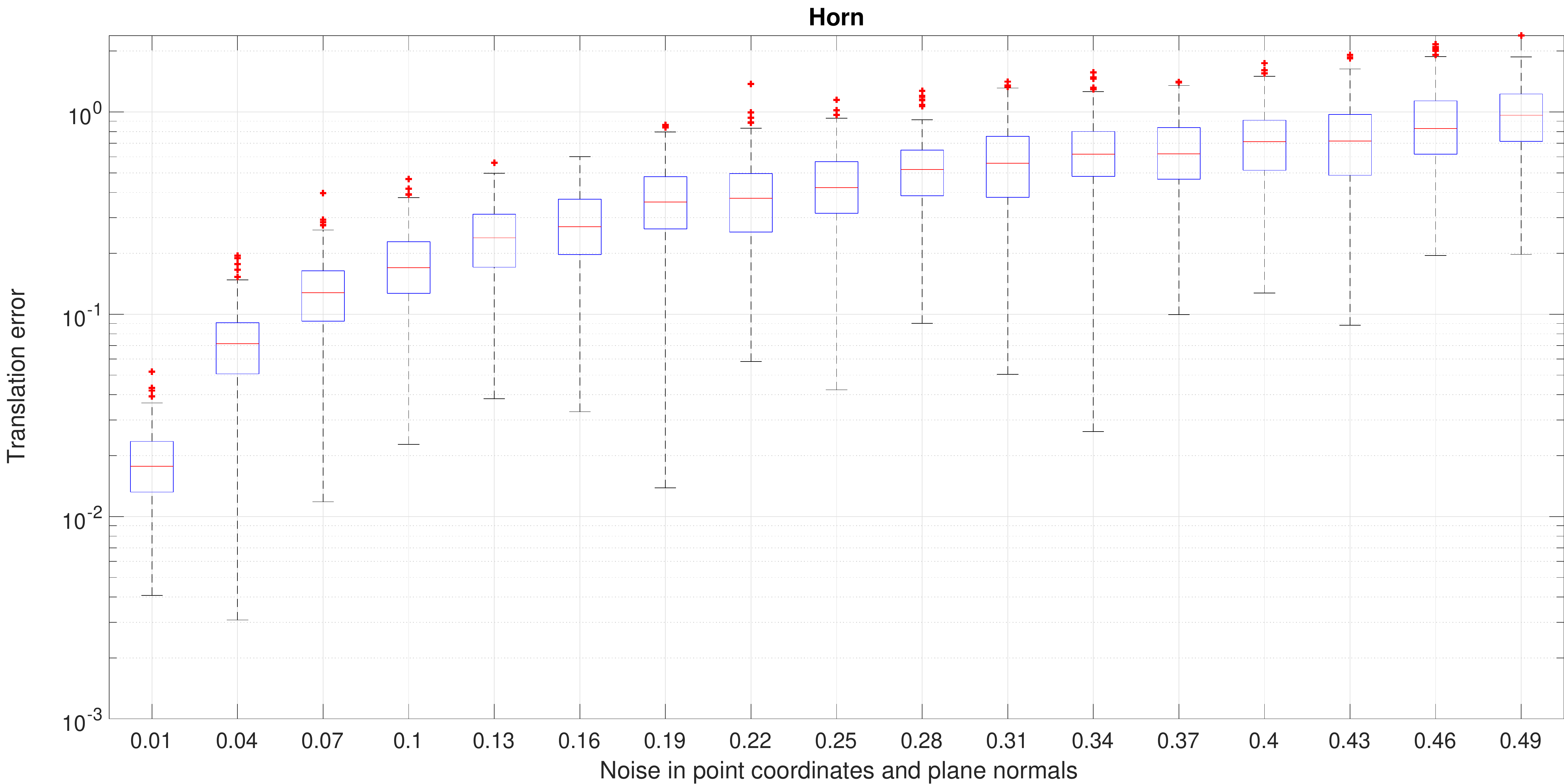}}
	\\
	\subfigure[SO(3) error for Horn's only]{\includegraphics[width=0.49\columnwidth]{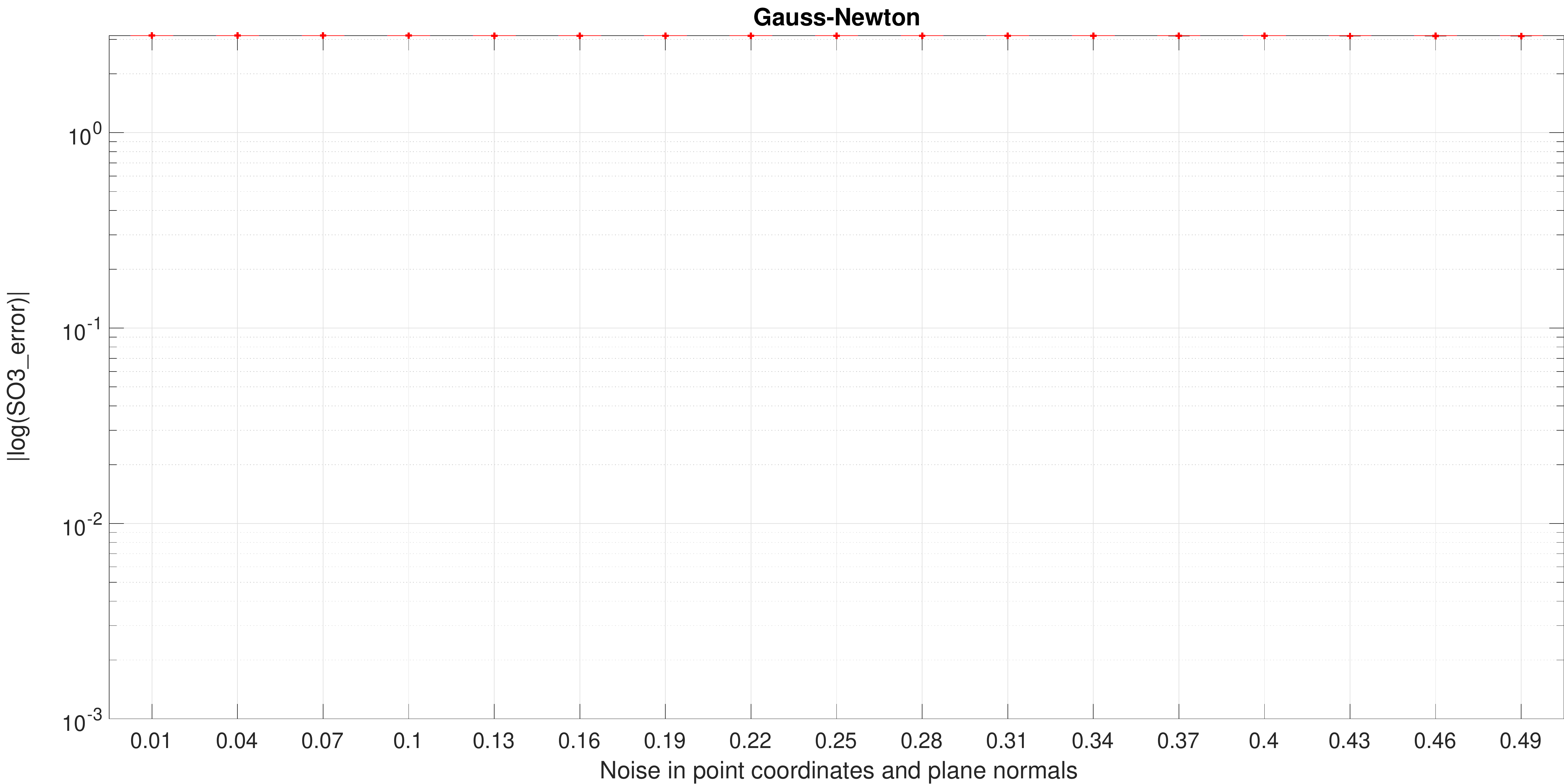}}
	\subfigure[Translation error for OLAE only]{\includegraphics[width=0.49\columnwidth]{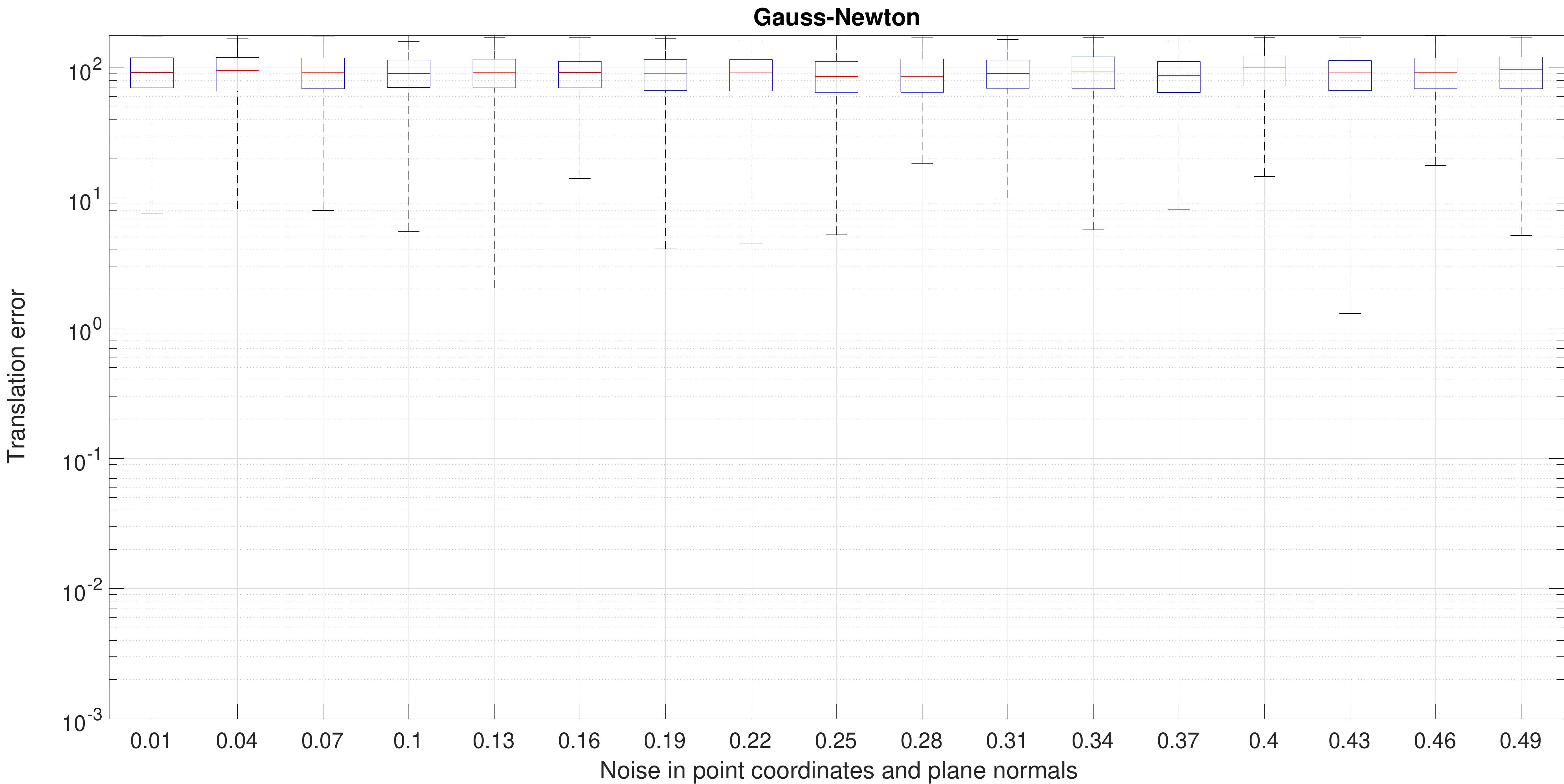}}
	\caption{Errors for the optimal transformation estimated by three different methods (OLAE, Horn's optimal quaternion, Gauss-Newton)
		for 1 point-to-point and 100 plane-to-plane correspondences. The horizontal axis represents the standard deviation of the points and planes noise.}
	\label{fig:results.pt1.pl100}
\end{figure*}

\begin{figure*}
	\centering
	\subfigure[SO(3) error ]{\includegraphics[width=0.49\columnwidth]{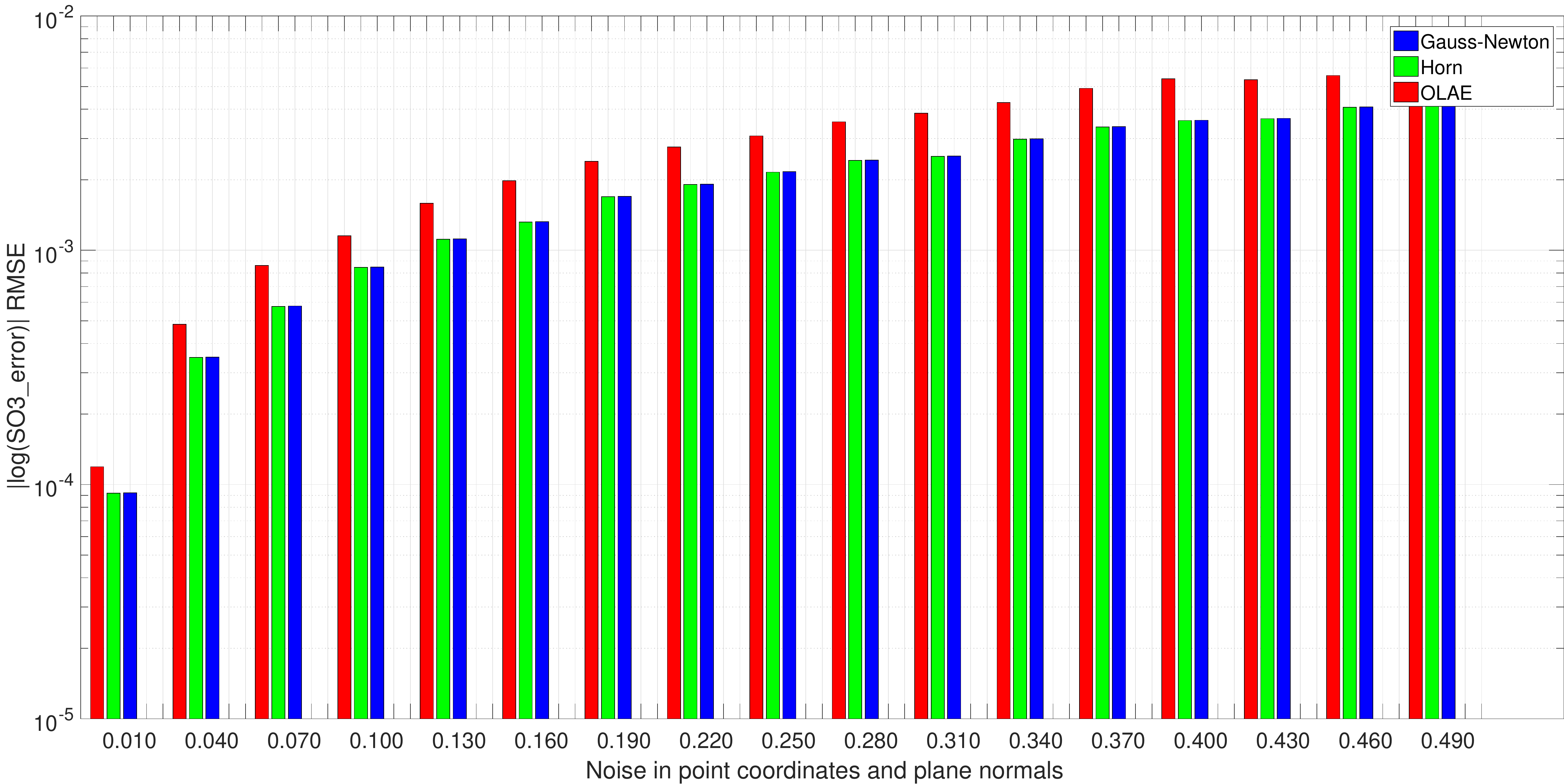}}
	\subfigure[Translation error]{\includegraphics[width=0.49\columnwidth]{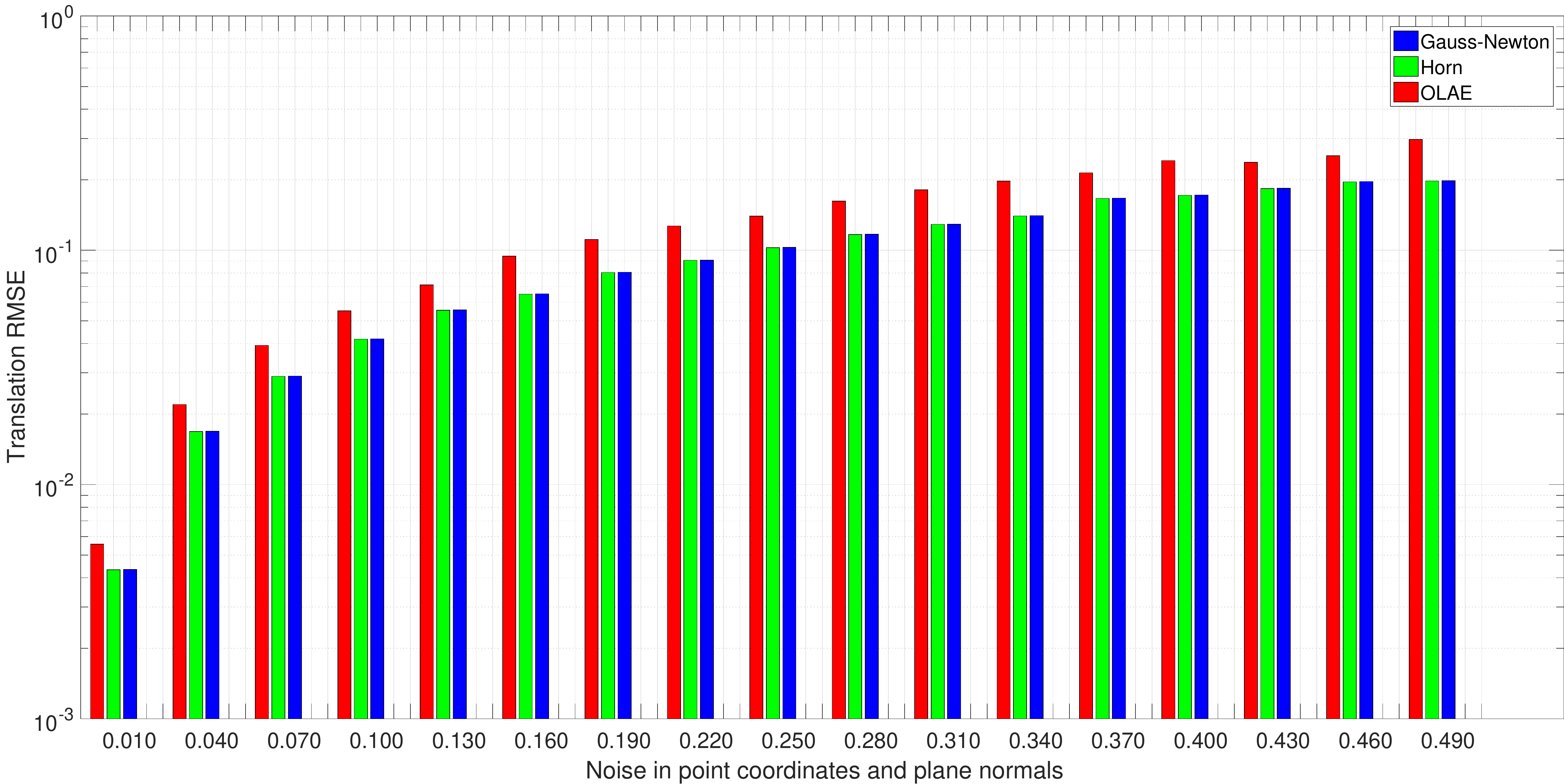}}
	\\
	\subfigure[SO(3) error for OLAE only]{\includegraphics[width=0.49\columnwidth]{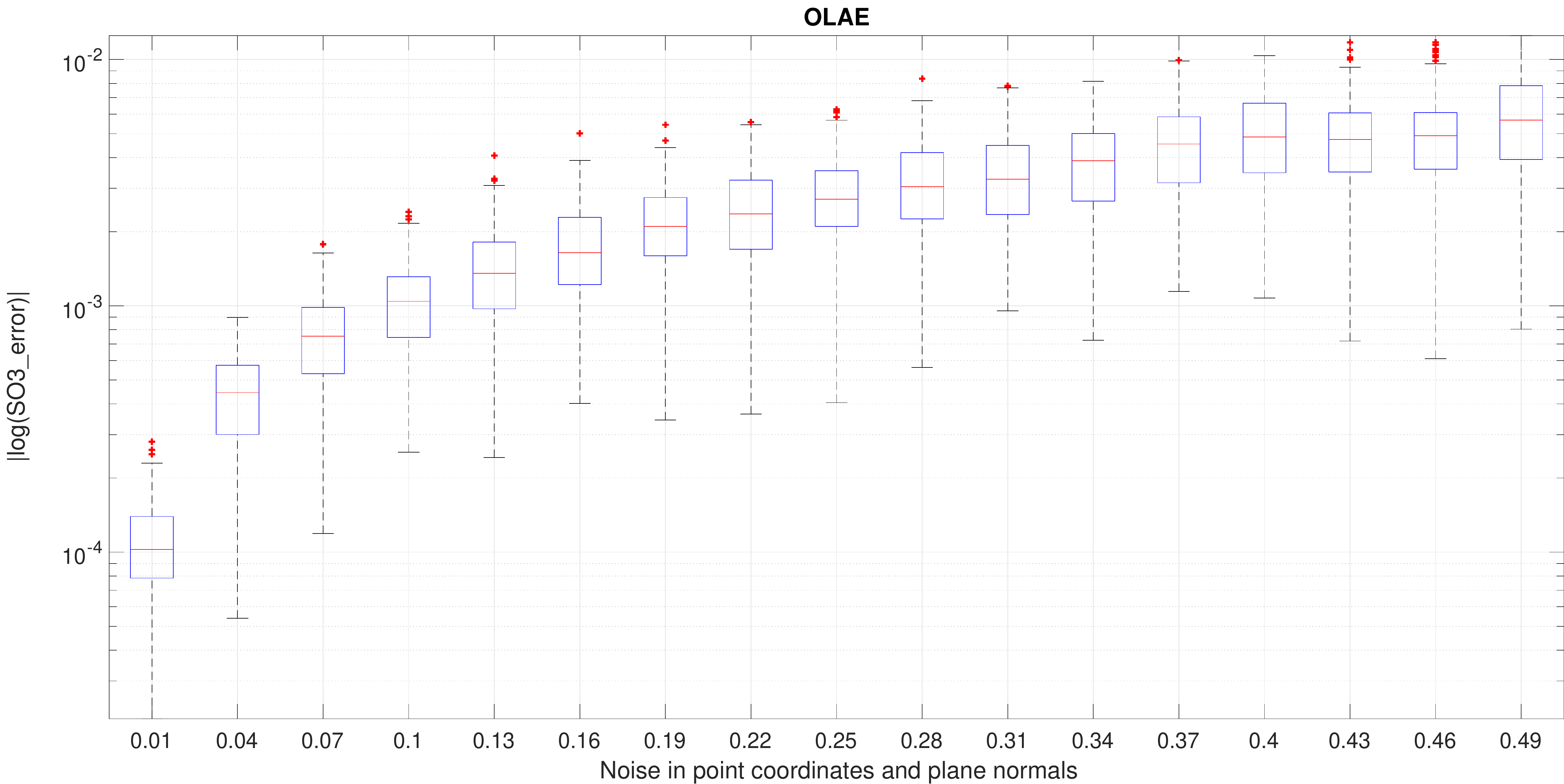}}
	\subfigure[Translation error for OLAE only]{\includegraphics[width=0.49\columnwidth]{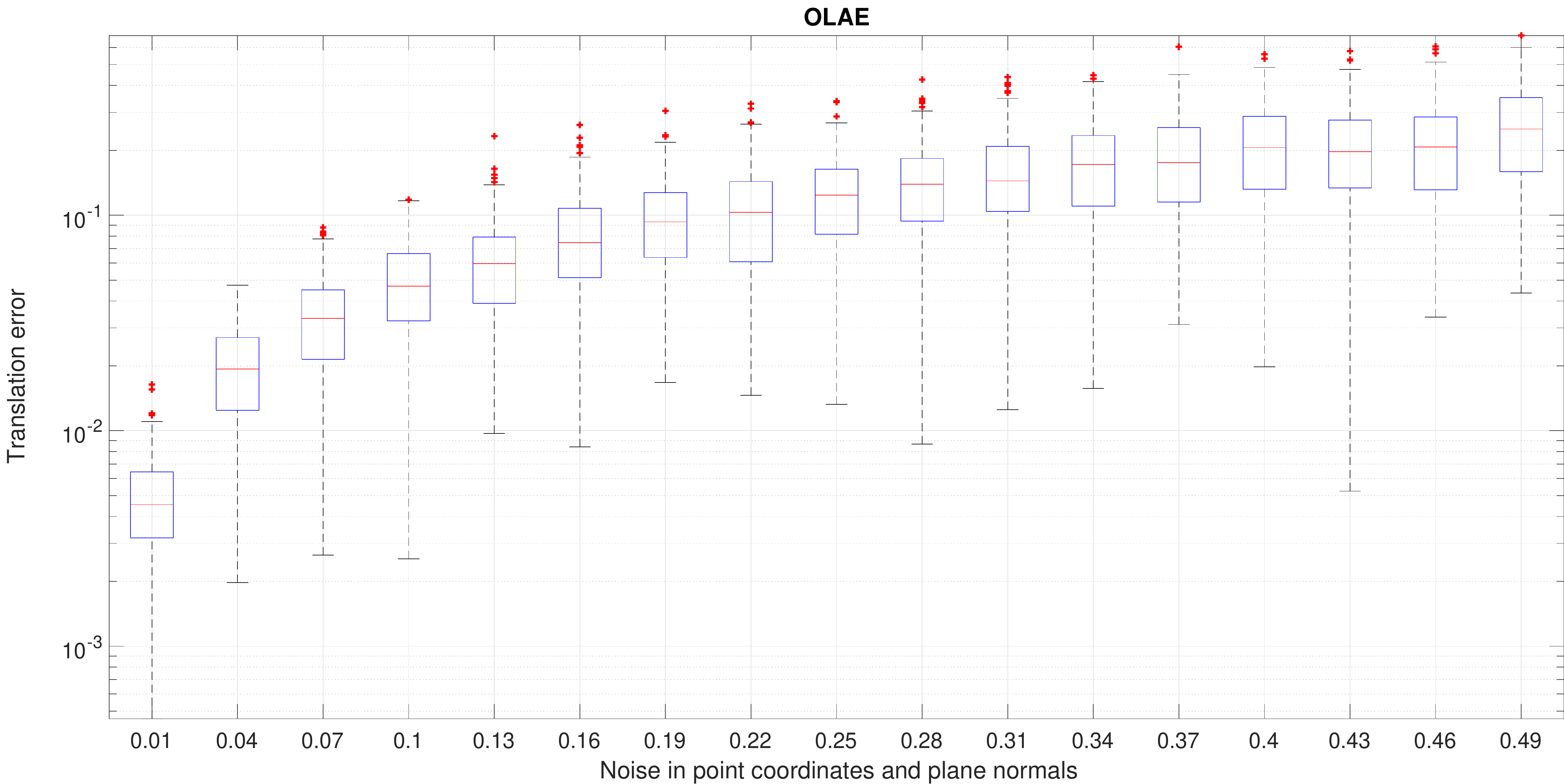}}
	\\
	\subfigure[SO(3) error for Horn's only]{\includegraphics[width=0.49\columnwidth]{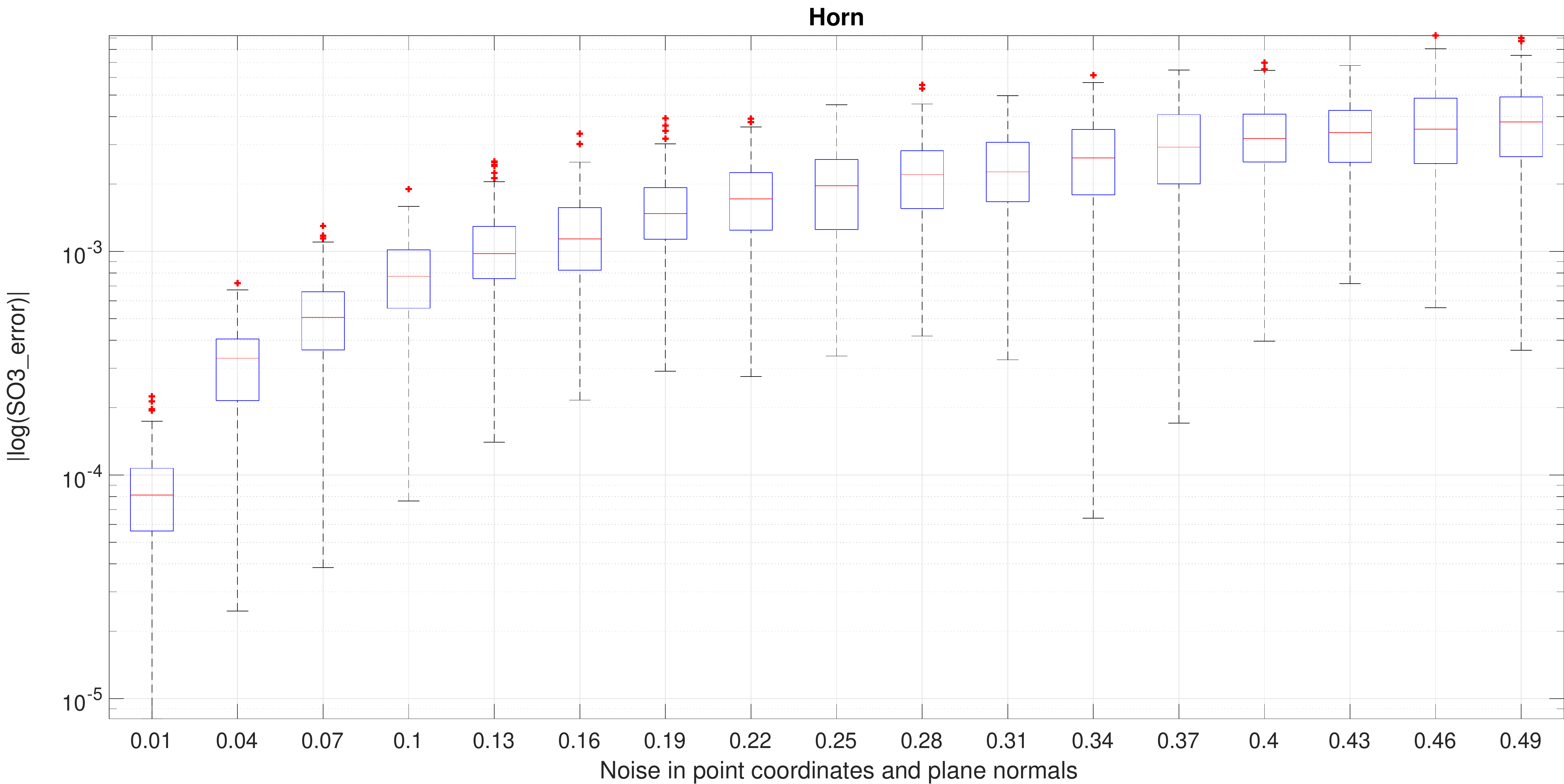}}
	\subfigure[Translation error for OLAE only]{\includegraphics[width=0.49\columnwidth]{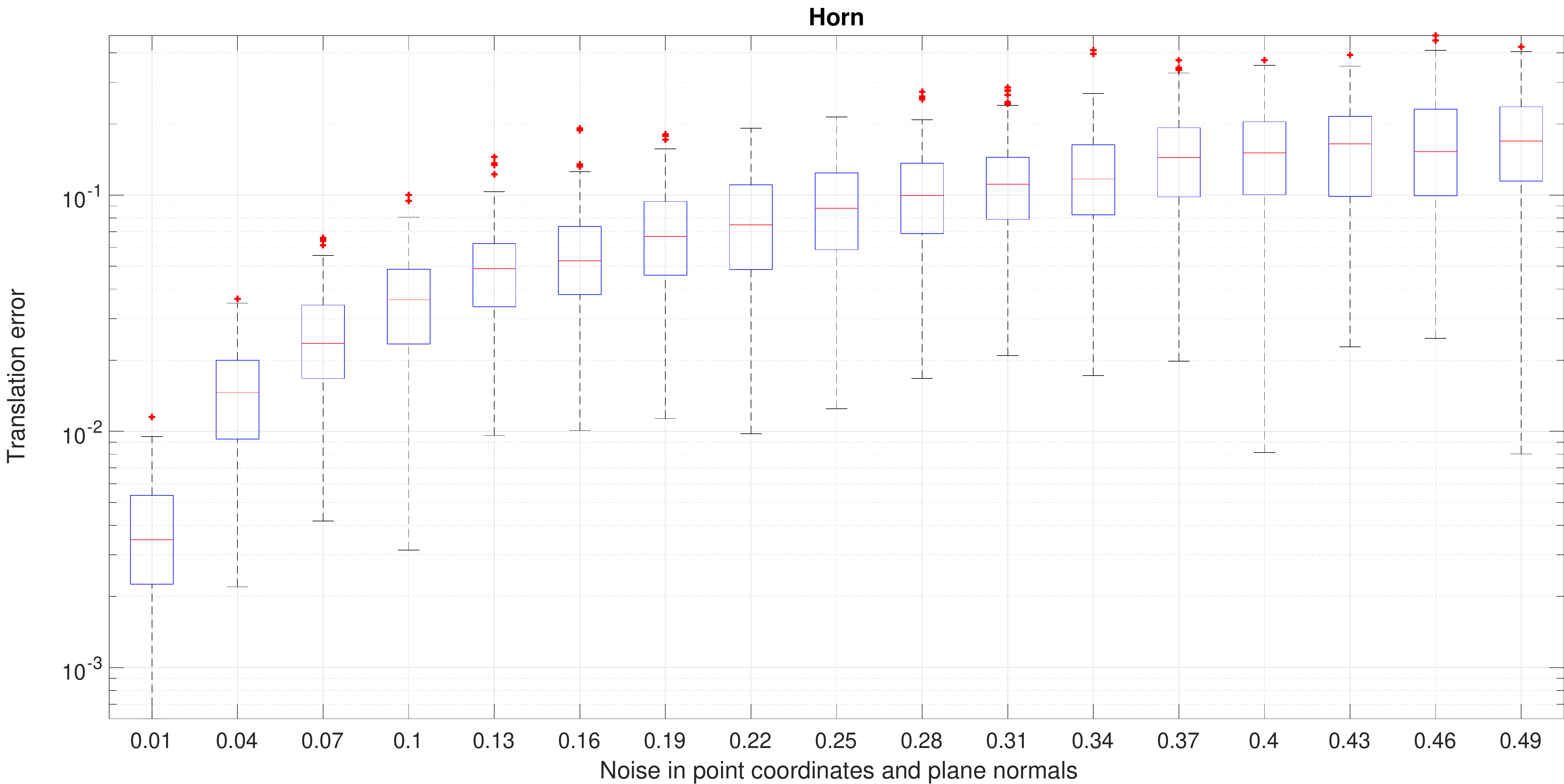}}
	\\
	\subfigure[SO(3) error for Horn's only]{\includegraphics[width=0.49\columnwidth]{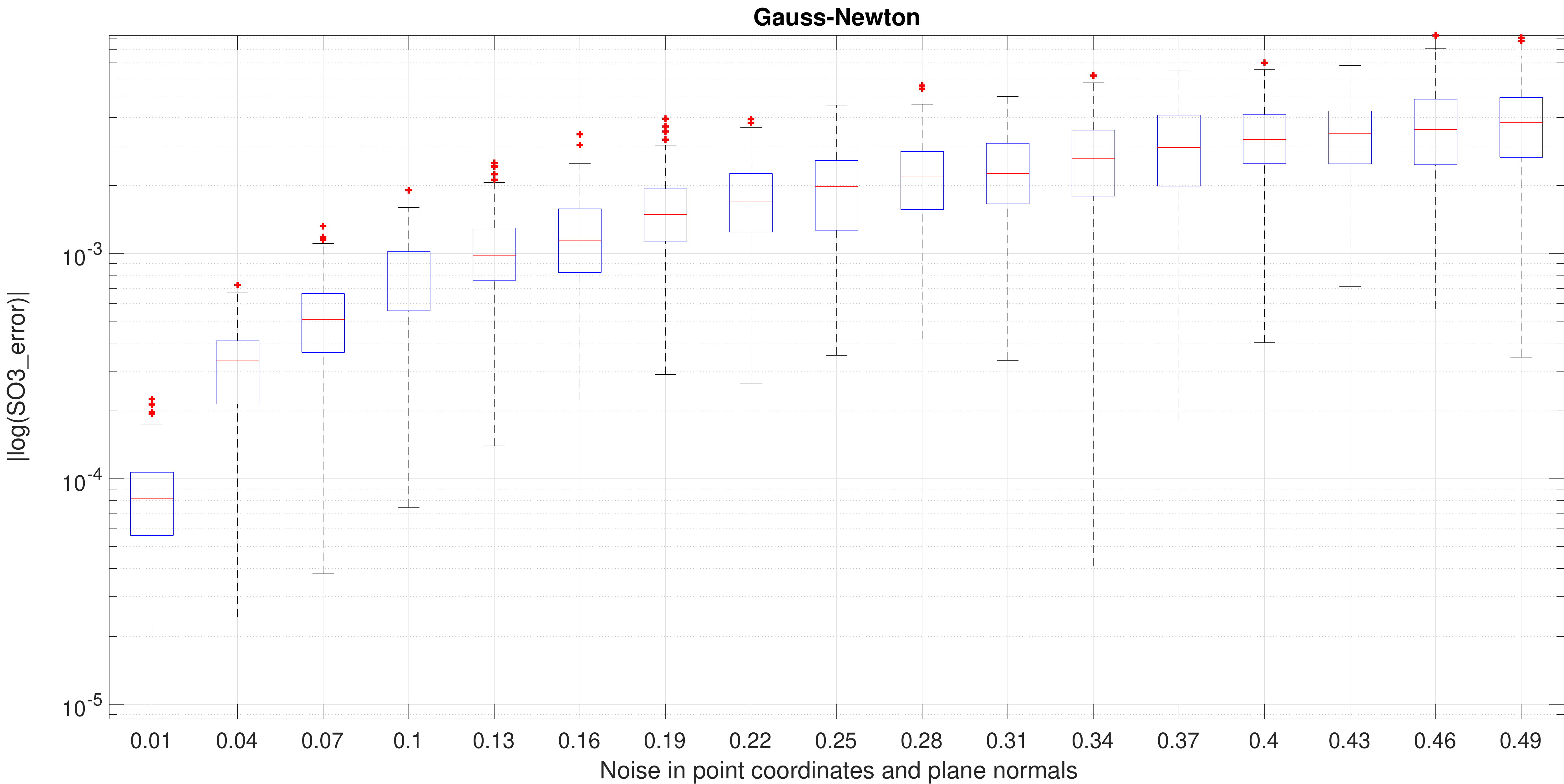}}
	\subfigure[Translation error for OLAE only]{\includegraphics[width=0.49\columnwidth]{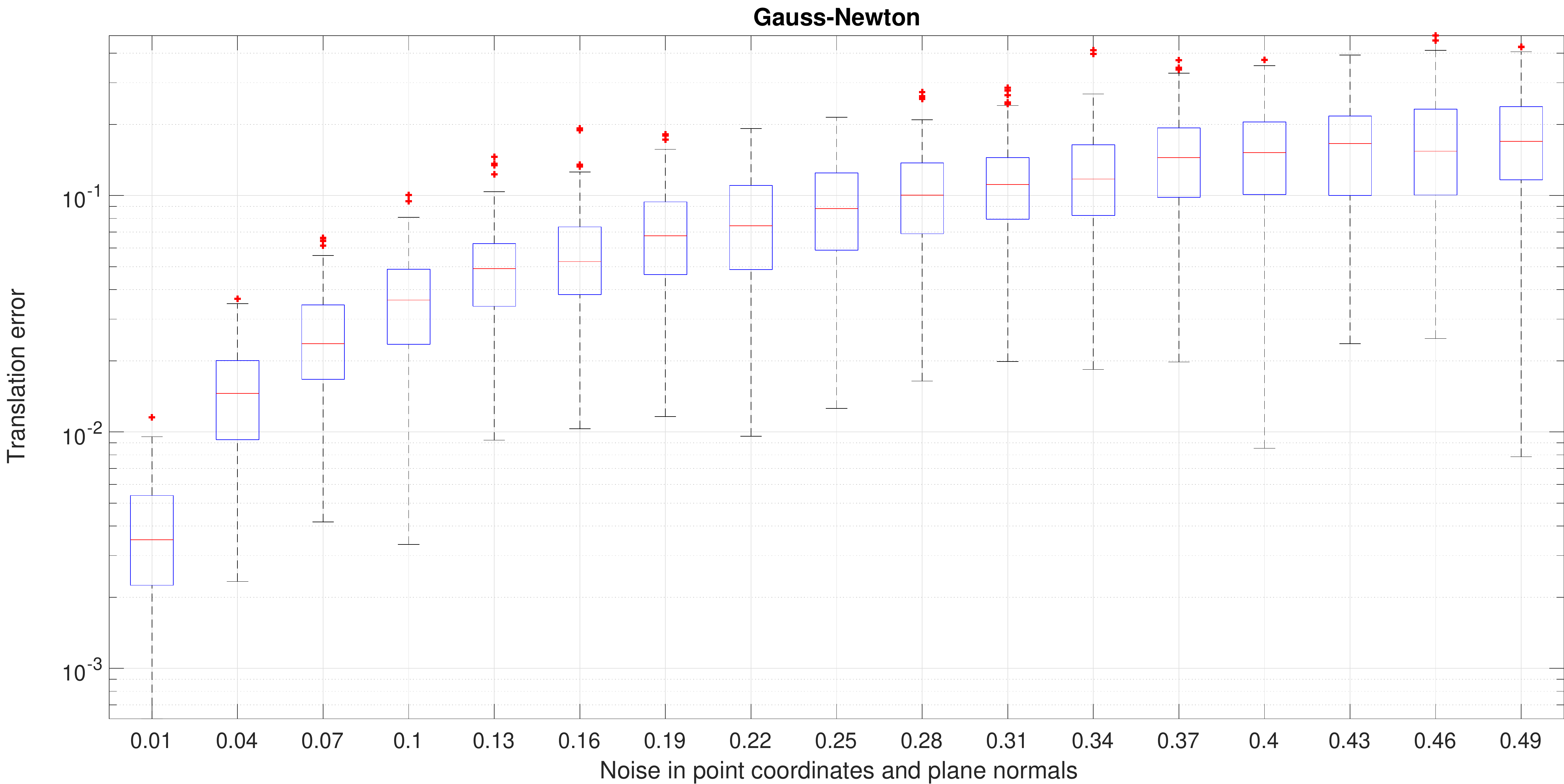}}
	\caption{Errors for the optimal transformation estimated by three different methods (OLAE, Horn's optimal quaternion, Gauss-Newton)
		for 100 point-to-point and 100 plane-to-plane correspondences. The horizontal axis represents the standard deviation of the points and planes noise.}
	\label{fig:results.pt100.pl100}
\end{figure*}

% ================================
\section{Experimental results}
\label{sect:experiments}
% ================================

Next we describe different numerical simulations aimed at evaluating the algorithms proposed in former sections.
The implementation used to obtain these results has been released as open source for the sake of reproducibility.

% ------------------------------------------
\subsection{Sensitivity to noise}

To benchmark the performance of the implemented methods against noise, sets of points and plane centroids are randomly drawn following an uniformly distributed in the cube (0,0,0)-(50,50,50), and then transformed following a random SE(3) pose which is then estimated from the three methods described in the text above.
Points in the transformed map are corrupted with additive Gaussian noise. Plane normals are also rotated following random SO(3) rotations whose rotation angle also follows a Gaussian distribution. Errors are evaluated in two parts: rotation error with respect to the ground truth is measured as the norm of the matrix logarithm of the rotation error, while translation is measured using Euclidean distances. All the experiments have been repeated 1000 times to obtain significant statistics.

Figure~\ref{fig:results.pt100.pl0} shows the statistical results 
for 100 paired points, while 
Figure~\ref{fig:results.pt1.pl100} 
shows the results for one point and 100 plane pairs, 
and finally Figure~\ref{fig:results.pt100.pl100} 
illustrates the case for 100 points and 100 planes.
No outliers are introduced in this test set.

% ------------------------------------------
\subsection{Absolute outlier rejection}
\label{sect:results.outlier}

\begin{figure}[h]
\centering
\subfigure[Attitude error]{\includegraphics[width=0.8\columnwidth]{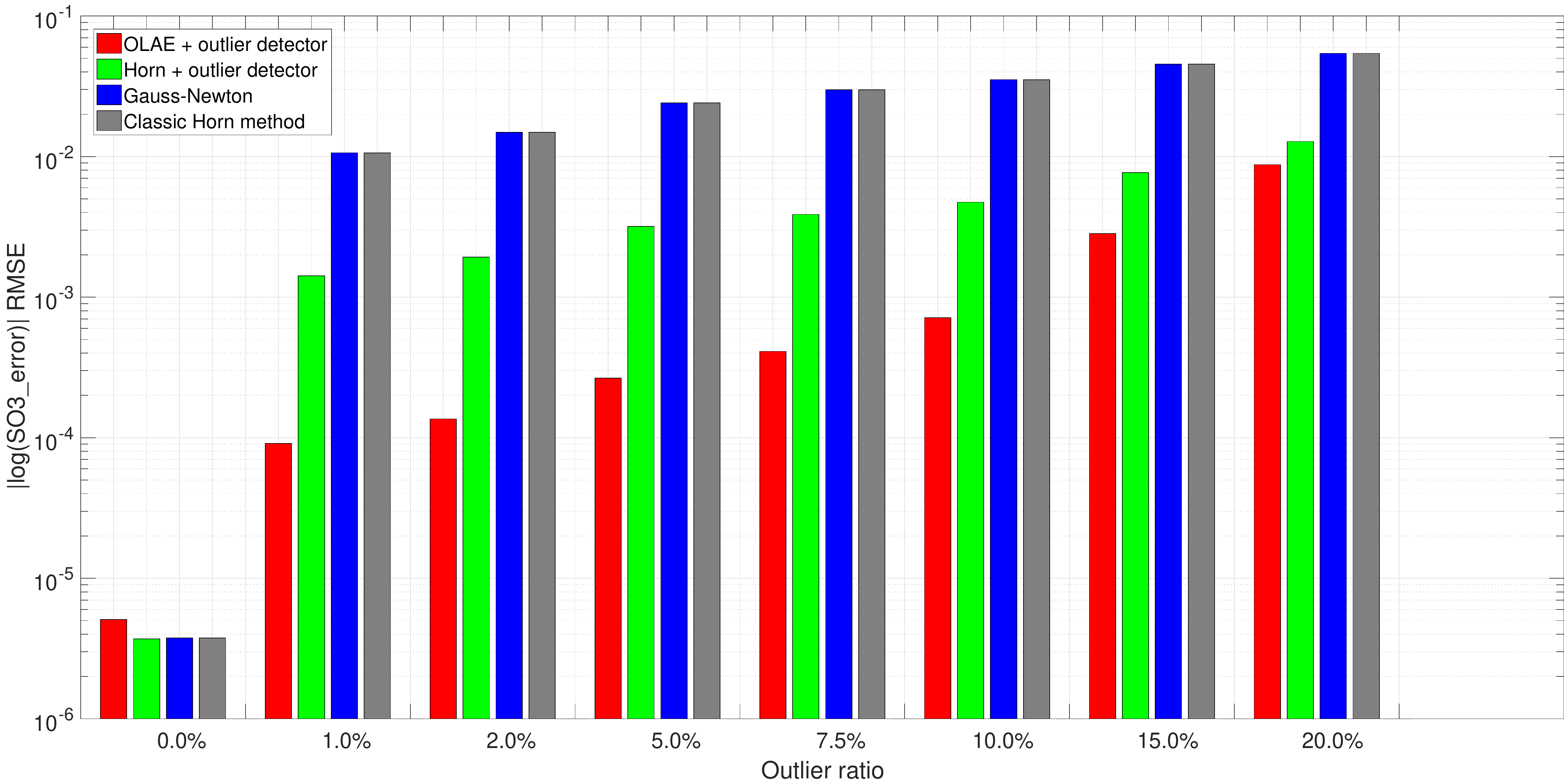}}
\\
\subfigure[Translation error]{\includegraphics[width=0.8\columnwidth]{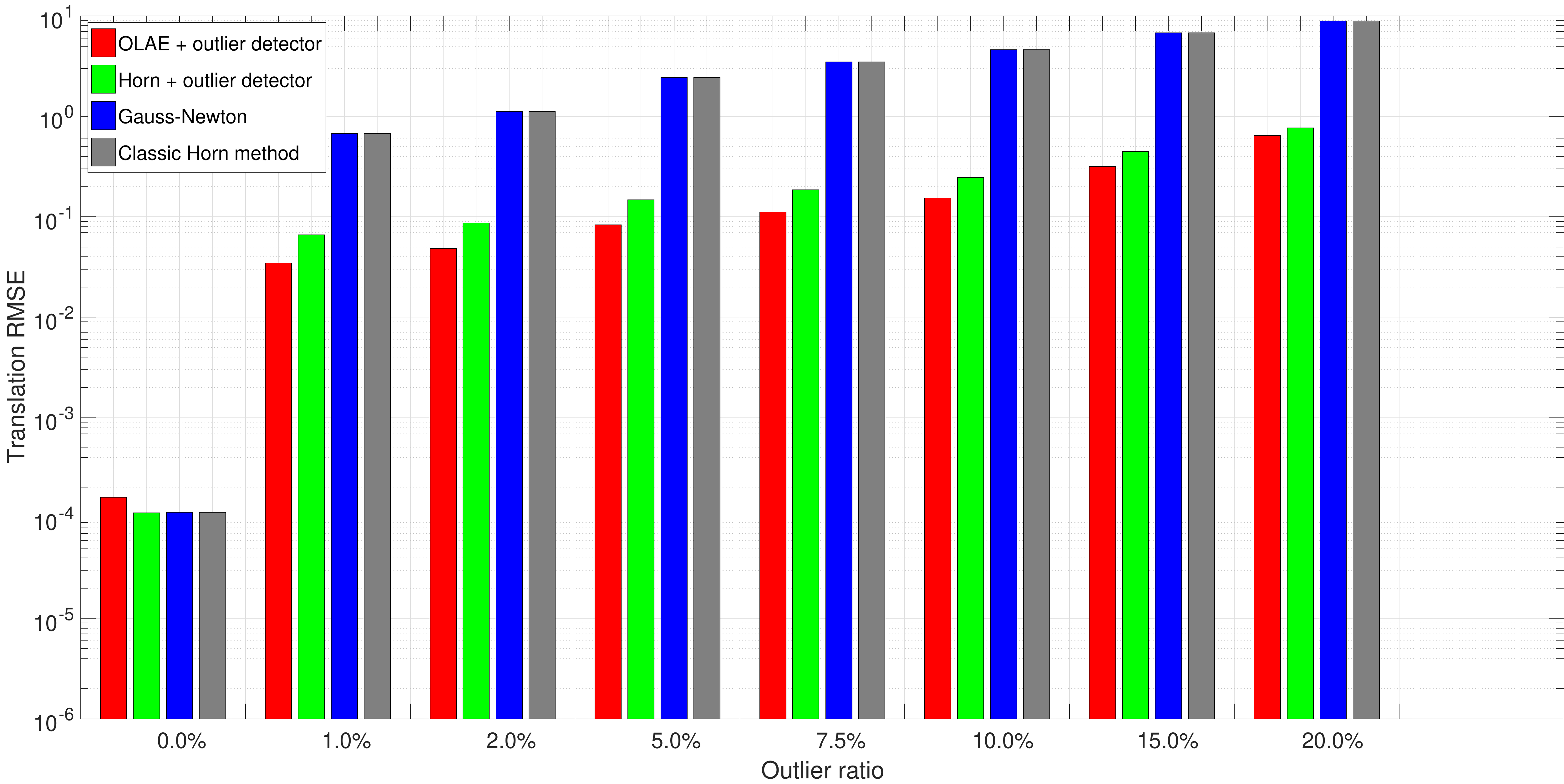}}
\caption{Optimal transformation error statistics with and without the scale-based absolute outlier rejection.}
\label{fig:results.outlier.rejection}
\end{figure}

Next we evaluate the outlier rejection method based on the scale mismatch threshold, as 
described by Eq.~(\ref{eq:outlier.test}).
The outlier detector has been integrated into both, OLAE and Horn's solution, 
and the error achieved after removing outliers (those that were successfully detected)
is shown in Figure~\ref{fig:results.outlier.rejection}.
For comparison, we added the result of the Gauss-Newton and the original Horn's method, 
both of them without any outlier filter.
Here we used the threshold $s_t=0.2$.

% ------------------------------------------
\subsection{Robust loss cost}

\begin{figure}[h]
\centering
\includegraphics[width=0.8\columnwidth]{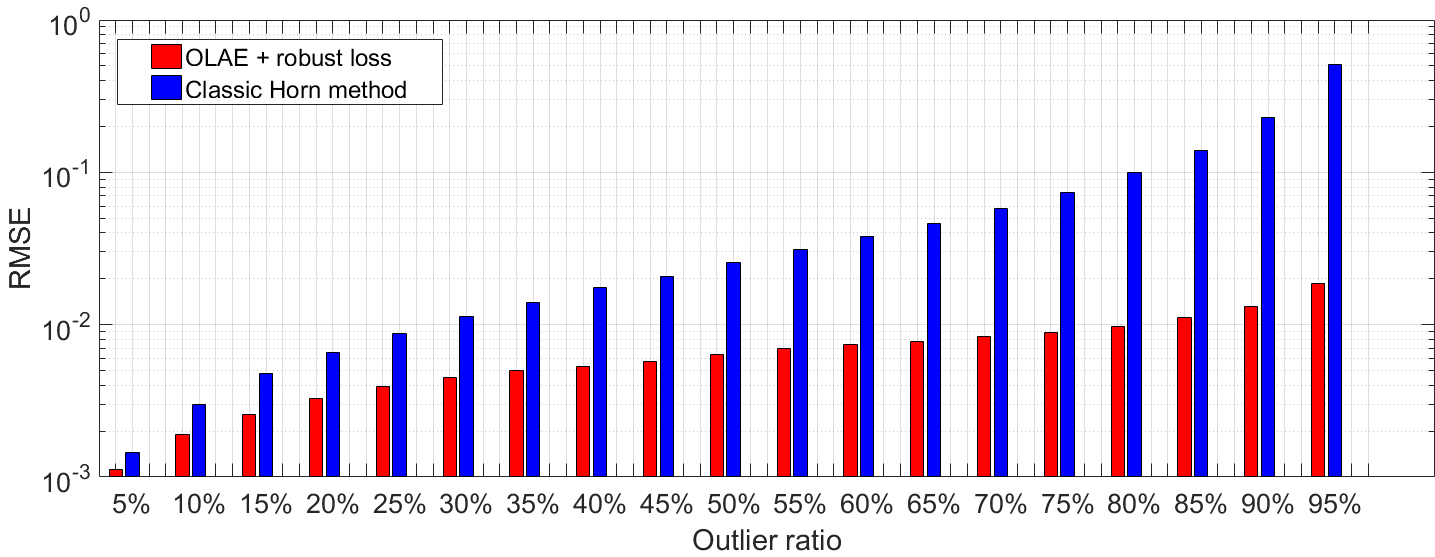}
\caption{Error for different outlier ratios with and without the 
	robust loss function to modify pair weights in the proposed ICP system. Note that this feature is only available when a reasonable initial estimate of the solution is known in advance.}
\label{fig:results.robust.loss}
\end{figure}

When a guess for the solution is available, e.g. in the final refining stages
of ICP, when the solution is near convergence, 
we can enable the additional robust loss weight factor mentioned in \S\ref{sect:complete.icp} to further 
mitigate the effect of outliers.
Figure~\ref{fig:results.robust.loss} shows a comparison of OLAE with the robust loss function and the classic Horn's solution (without loss function). 

% ------------------------------------------
\subsection{Complete ICP system}

\begin{figure}
	\centering
	\subfigure[SO(3) rmse]{\includegraphics[width=0.45\columnwidth]{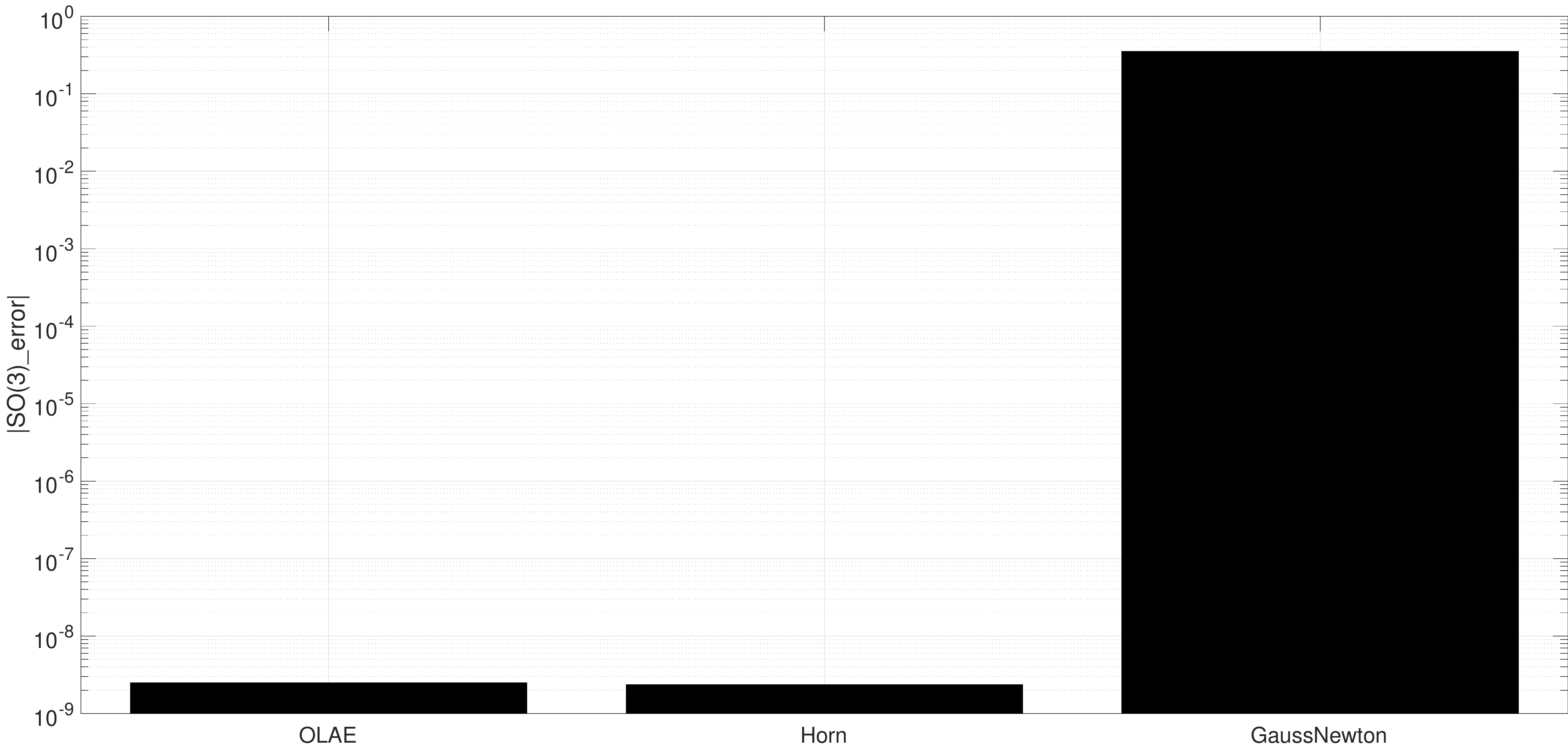}}
	\subfigure[Translation rmse]{\includegraphics[width=0.45\columnwidth]{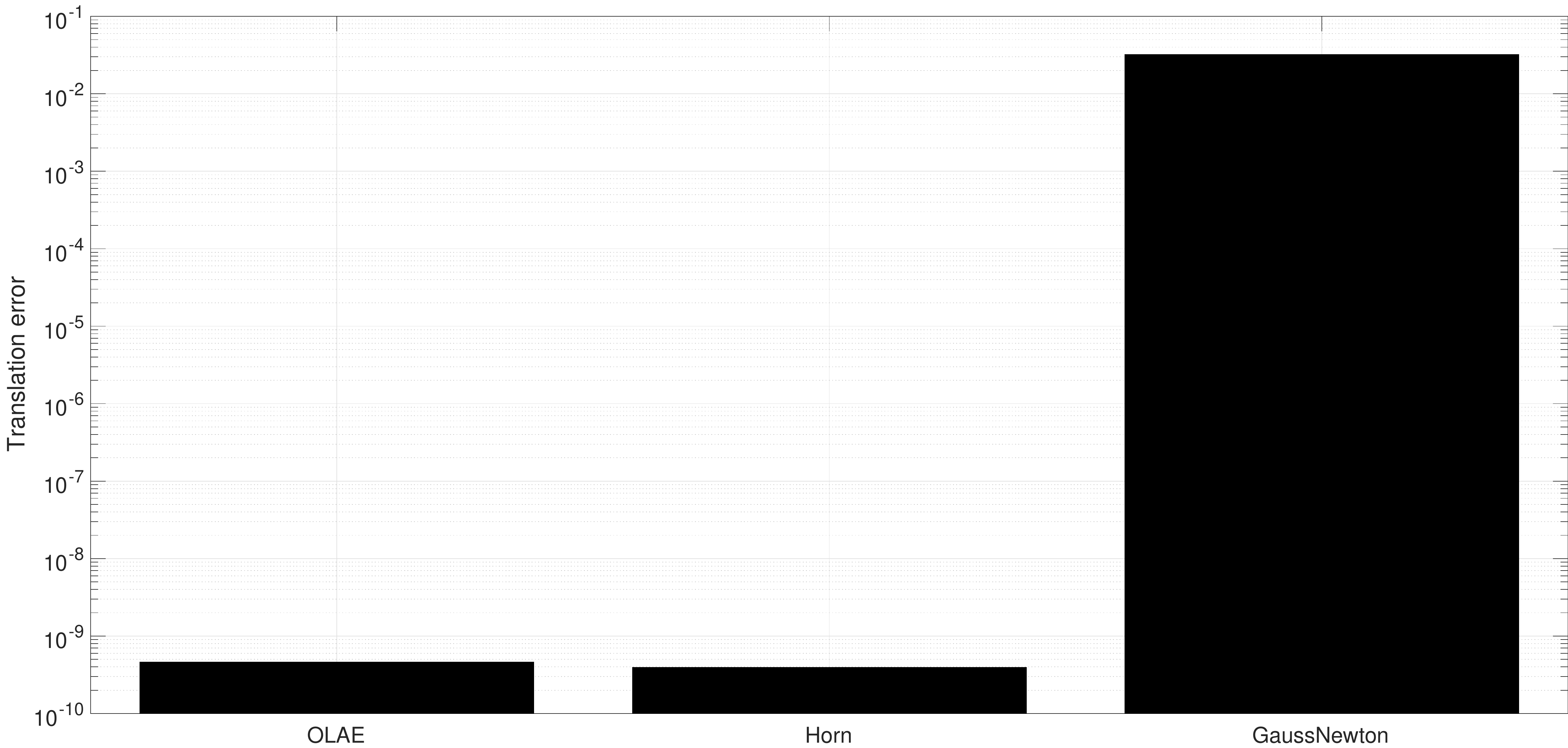}}
	\subfigure[Computation time]{\includegraphics[width=0.45\columnwidth]{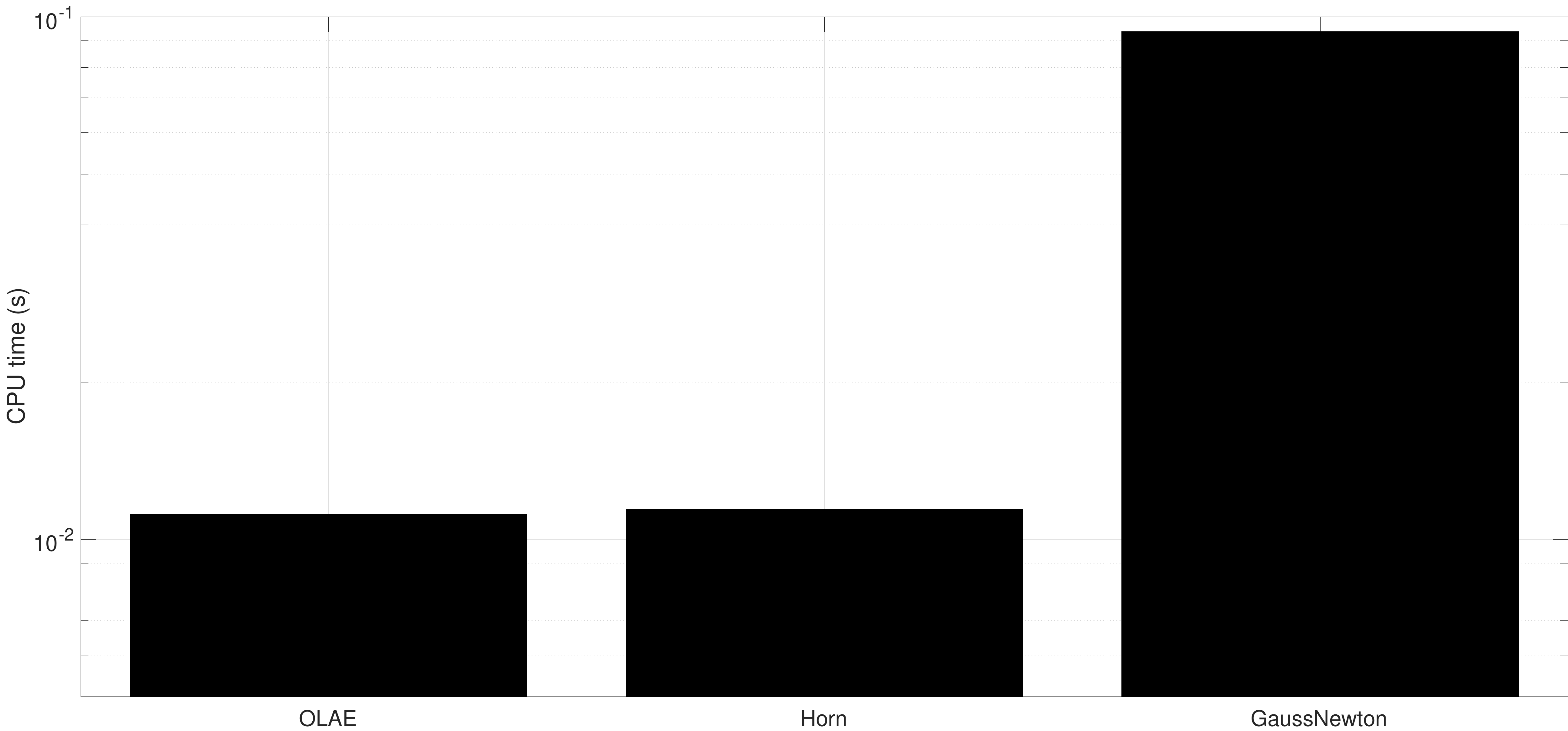}}
	\caption{Results for the entire ICP system, when using each of the different algorithms as optimal transformation solver at it core.}
	\label{fig:results.icp}
\end{figure}

In order to test the complete ICP system described in \S\ref{sect:complete.icp}, 
we used 3D pointcloud models publicly available in the Stanford's dataset (\cite{curless1996volumetric}).
In particular, in this section we used the \emph{Bunny} dataset, downsampled to 1000 points, as a reference 
pointcloud. 
Then, a transformed pointcloud is generated by translating and rotating the model using a random SE(3) pose
with translation in X,Y, and Z, uniformly drawn in the range $[-0.25 b, 0.25 b]$, with $b$ the maximum length of the model bounding box, 
and with random rotations built from values of yaw, pitch, and roll drawn from a uniform distribution in the range $[-20 deg, +20 deg]$. Note that, while the optimal transformation methods (OLAE, Horn's) are able to 
find a global optimal transformation without iterating, the \emph{association} stage required in ICP
limits the convergence volume of the state space; that explains the relatively small translations and rotations used in this test.
The whole process is repeated 10 times, and we measured the SO(3) rmse, the translation rmse, and CPU time, of the final estimation 
of the ICP algorithm when using each of the three different algorithms discussed above at its core while solving for optimal transformations.
The statistical results can be seen in Figure~\ref{fig:results.icp}. 

\newpage

% ================================
\section{Discussion and conclusions} 
\label{sec:conclusion}
% ================================

We have presented a methodology to allow point, line, and plane features to be integrated into Horn's method 
and into OLAE, 
which originally only supported point and vector observations, respectively.

It is worth mentioning that some information is lost due to the vector normalization stage in Eq.~(\ref{eq:olae.vec.norm}) when using OLAE (but not in Horn's method), and this has an impact 
in the solution accuracy: when most correspondences are points, attitude-base methods (i.e. OLAE)
performed slightly worse than Horn's method, which in turn considers the full scale of relative point coordinates.
From the statistical results, it can be concluded that 
OLAE performs identical to Horn's method when most features are vector-like (i.e. planes or lines). 
OLAE is faster than Horn's method only when working with a reduced number of correspondences (roughly less than 10 primitive pairings), thus it should be the preferred choice only for applications that require the fast evaluation of small sets, e.g. inside a RANSAC loop.

Future works include methods to avoid relying on a centroid. This has revealed as a weak point 
of the Horn's classic solution, 
which may be mitigated by using relative 
vectors for different graph topologies, as recently proposed in \cite{yang2019polynomial}.

%% Use plainnat to work nicely with natbib. 
%\bibliographystyle{unsrtnat}
%\bibliography{p2p2_robust_pc_align}

\end{document}